\documentclass[10pt,twocolumn,letterpaper]{article}

\usepackage{iccv}
\usepackage{times}
\usepackage{epsfig}
\usepackage{graphicx}
\usepackage{amsmath}
\usepackage{amssymb}
\usepackage{array}
\usepackage{booktabs}
\usepackage{enumitem}
\usepackage{subfigure}
\usepackage{capt-of}
\usepackage{xcolor}
\usepackage{cuted}
\usepackage{multirow}
\usepackage{caption}
\usepackage{color}

\usepackage[pagebackref=true,breaklinks=true,letterpaper=true,colorlinks,bookmarks=false]{hyperref}

\iccvfinalcopy 

\def\iccvPaperID{3500} 

\ificcvfinal\fi

\begin{document}

\title{SSH: A Self-Supervised Framework for Image Harmonization}

\author{%
  Yifan Jiang$^{\dagger}$, He Zhang$^{\ddagger}$, Jianming Zhang$^{\ddagger}$, Yilin Wang$^{\ddagger}$, Zhe Lin$^{\ddagger}$, Kalyan Sunkavalli$^{\ddagger}$, \\Simon Chen$^{\ddagger}$, Sohrab Amirghodsi$^{\ddagger}$, Sarah Kong$^{\ddagger}$, Zhangyang Wang$^{\dagger}$\\
  $^{\dagger}$The University of Texas at Austin \,\,\,
  $^{\ddagger}$Adobe Inc.\\
  $^{\dagger}$\texttt{\small{\{yifanjiang97,atlaswang\}}@utexas.edu}, \\ 
  $^{\ddagger}$\texttt{\small{\{hezhan,jianmzha,yilwang,zlin,sunkaval,sichen,tamirgho,sakong\}}@adobe.com} \\
}

\twocolumn[{%
\renewcommand\twocolumn[1][]{#1}%
\maketitle
\begin{center}
 \centering
 \small
 \setlength{\tabcolsep}{0.0pt}
 \vspace{-1em}
 \begin{tabular}{cccccc}
     \includegraphics[width=0.16\textwidth]{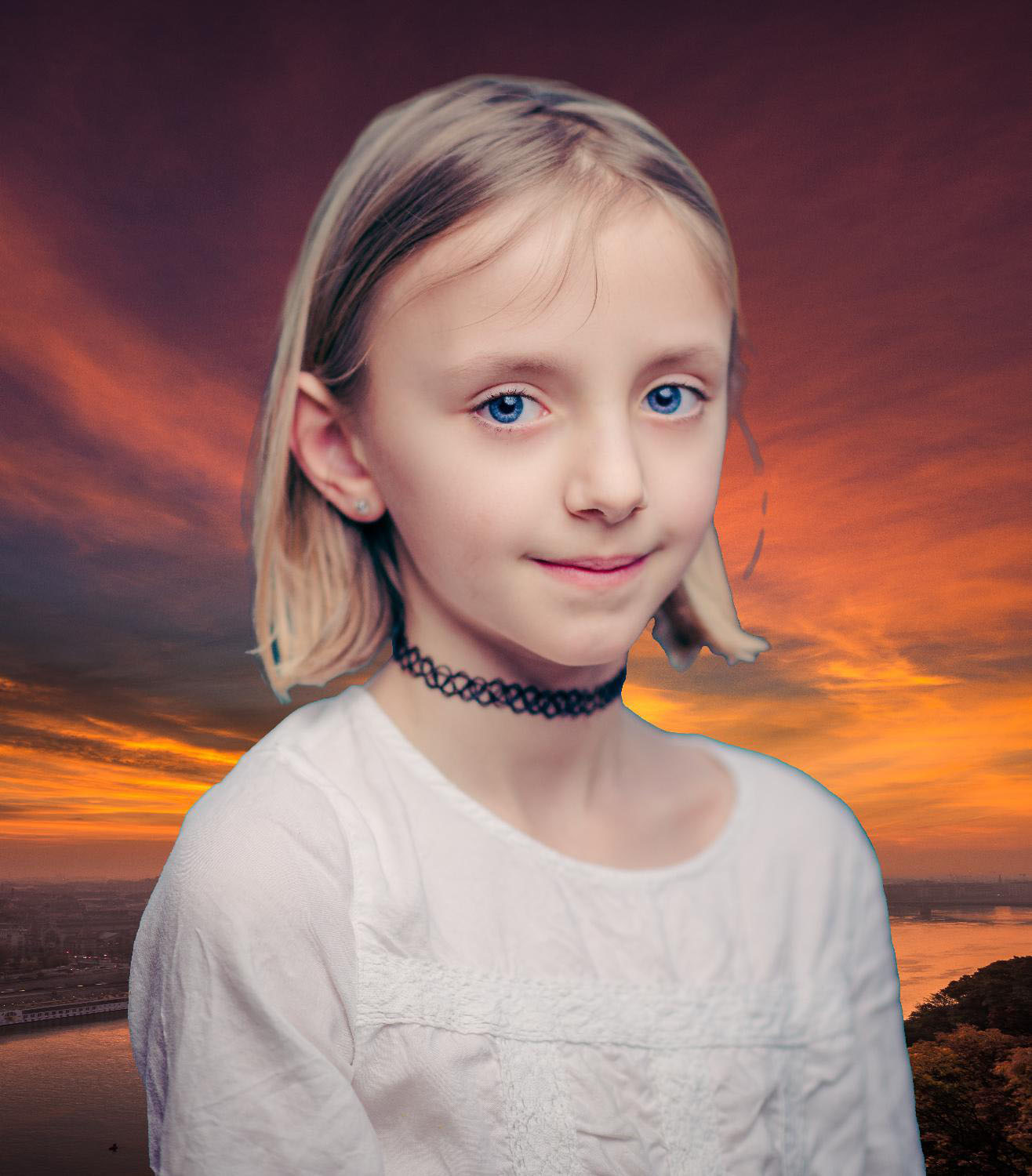} & \hspace{-0.1em}
     \includegraphics[width=0.162\textwidth]{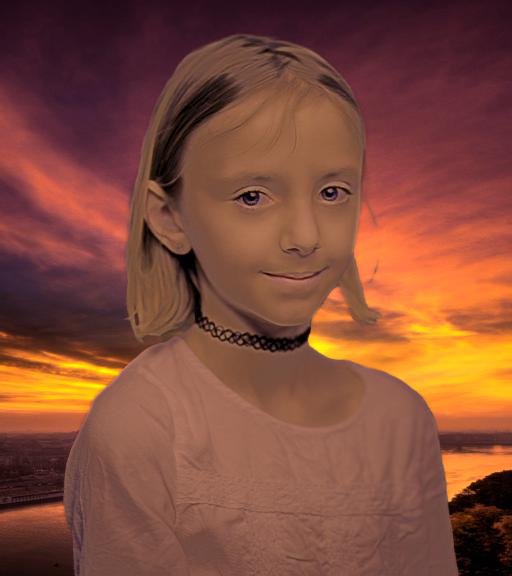} & \hspace{-0.1em}
     \includegraphics[width=0.16\textwidth]{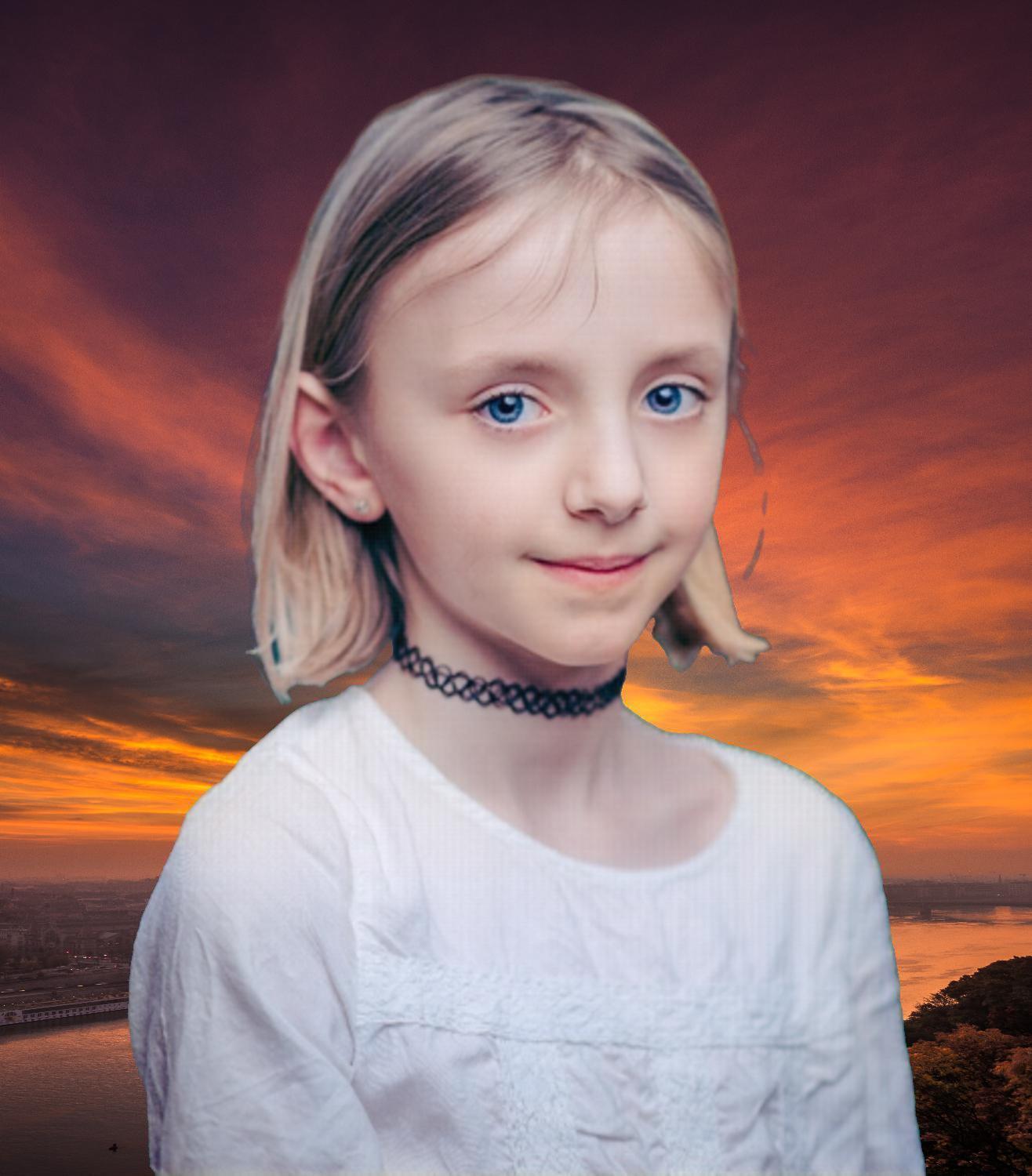} & \hspace{-0.1em}
     \includegraphics[width=0.16\textwidth]{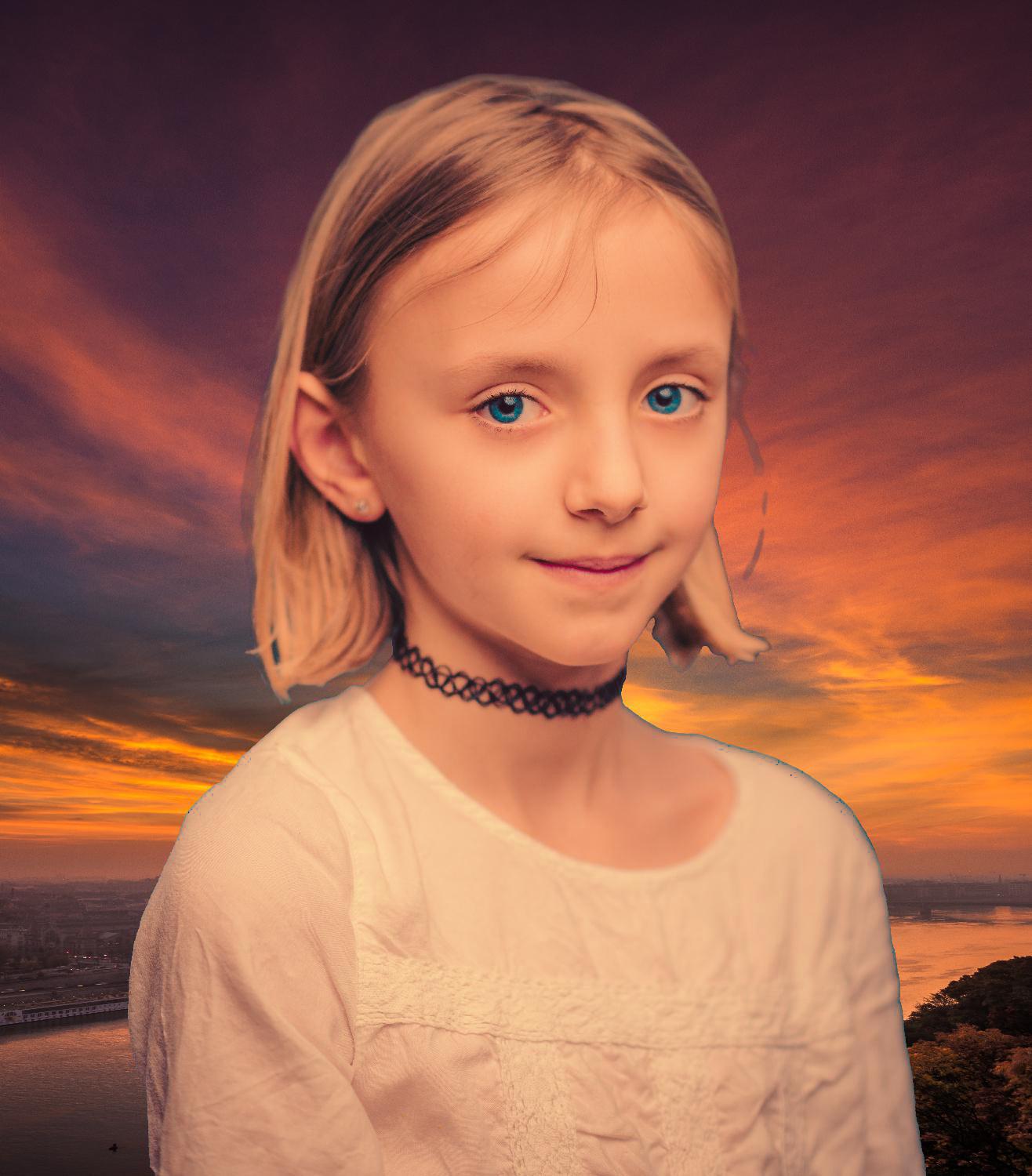} & \hspace{-0.1em}
     \includegraphics[width=0.16\textwidth]{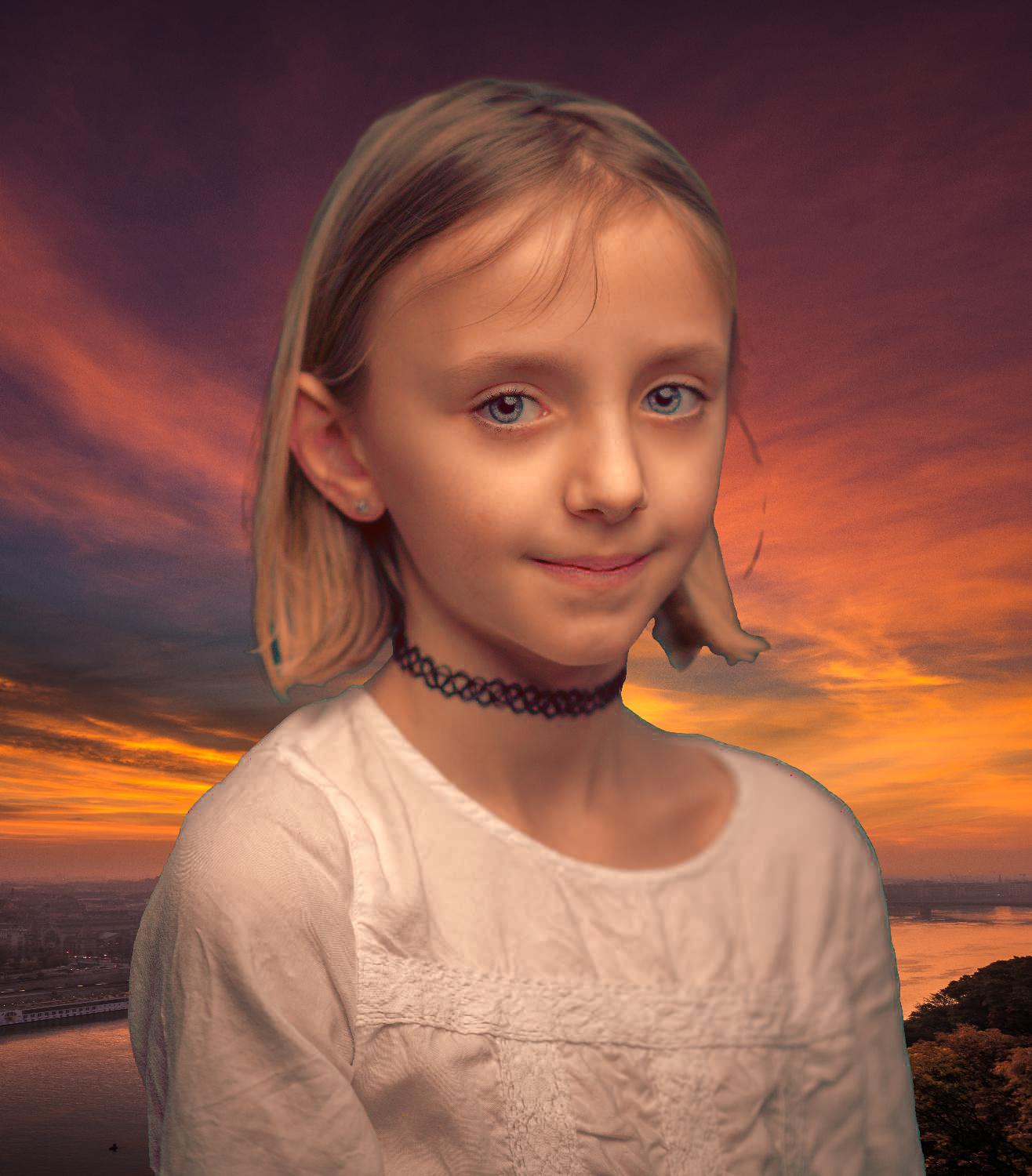} & \hspace{-0.1em}
     \includegraphics[width=0.16\textwidth]{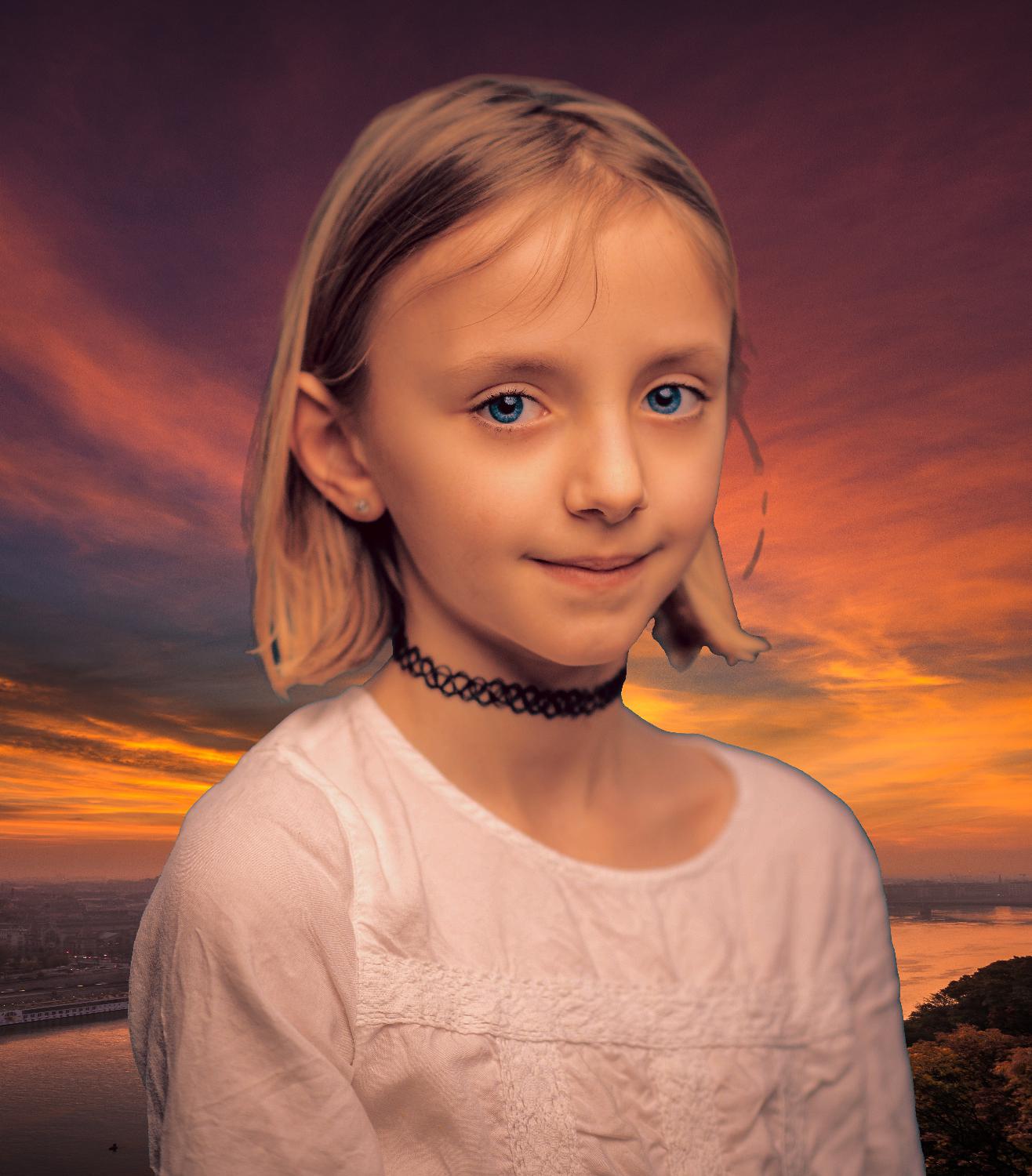}
     \\
     DC &
     $WCT^{2}$ \cite{yoo2019photorealistic} &
     DIH \cite{tsai2017deep} &
     $S^2$AM \cite{cun2020improving} &
     DoveNet \cite{cong2020dovenet}&
     SSH (ours) 
 \end{tabular}
\vspace{-1em}
 \captionof{figure}{A visual comparison between the image harmonization results of the direct composition (DC), the current state-of-the-art methods, and the proposed SSH. Best viewed in color and zoomed in.}
 \label{fig:Teaser}
\end{center}%
}]

\begin{abstract}
\vspace{-1em}
Image harmonization aims to improve the quality of image compositing by matching the ``appearance" (\eg, color tone, brightness and contrast) between foreground and background images. However, collecting large-scale annotated datasets for this task requires complex professional retouching. 
Instead, we propose a novel Self-Supervised Harmonization framework (SSH) that can be trained using just ``free" natural images without being edited. 
We reformulate the image harmonization problem from a representation fusion perspective, which separately processes the foreground and background examples, to address the background occlusion issue.
This framework design allows for a dual data augmentation method, where diverse [foreground, background, pseudo GT] triplets can be generated by cropping an image with perturbations using 3D color lookup tables (LUTs).
In addition, we build a real-world harmonization dataset as carefully created by expert users, for evaluation and benchmarking purposes.
Our results show that the proposed self-supervised method outperforms previous state-of-the-art methods in terms of reference metrics, visual quality, and subject user study. Code and dataset are available at \url{https://github.com/VITA-Group/SSHarmonization}.
\vspace{-1em}
\end{abstract}

\vspace{-0.5em}
\section{Introduction}
\vspace{-0.5em}
Image harmonization is a crucial step in image compositing that aims at adjusting (harmonizing) the \textit{appearance}---\eg, the color, saturation, brightness and contrast---of a foreground object to better match the background image so that the resulting composite is more realistic. For example, a subject captured under sunlight looks different from one on a cloudy day and its appearance needs to be edited when composited into a cloudy scene.


\begin{figure*}
\centering   
\includegraphics[width=0.99\linewidth]{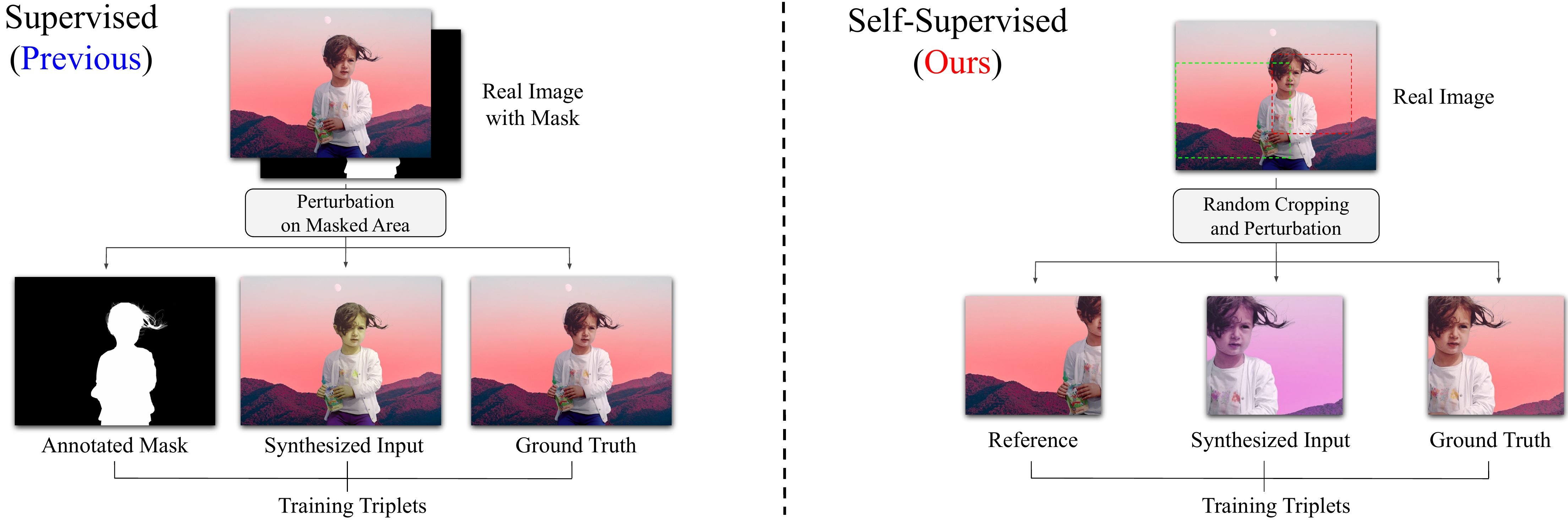}
\caption{Comparison between previous supervised methods~\cite{cong2020dovenet,cun2020improving,tsai2017deep} and the proposed SSH. Unlike previous methods that demand annotated masks in the training process, our self-supervised framework requires \underline{NO mask} during training. \textbf{The annotated mask is only needed at testing for actually making the composed image. }} 
\label{fig:Difference}
\end{figure*}

Previous approaches tackle this issue by transferring the statistic information between the foreground and background regions, including color \cite{lalonde2007using,xue2012understanding} and texture \cite{sunkavalli2010multi}. More recently, \cite{tsai2017deep,cong2020dovenet,cun2020improving} train deep neural networks to address the image harmonization problem, necessitating the large-scale dataset of input-harmonized composite training pairs.
However, collecting a large-scale high-quality harmonization dataset, in general, requires tedious professional expert retouching.
Instead, existing methods~\cite{tsai2017deep,cong2020dovenet,cun2020improving} bypass this by selecting foreground objects in existing images, perturb their color to \emph{simulate} an unharmonized composite, and train the network to regress the original input image, as manifested in Fig.~\ref{fig:Difference} left.
While these approaches~\cite{tsai2017deep,cong2020dovenet,cun2020improving} are effective to an extent, they have several limitations: 
\begin{itemize}[leftmargin=*]
    \item \textbf{Limited ground truth paired data.} Collecting high-quality paired harmonization data is time-consuming and laborious. Even in the constrained case presented above, it requires an accurate mask of the foreground object in each image, shown as Fig.~\ref{fig:Difference} left.  
    \item \textbf{Background occlusion.} Due to synthesizing by naively composing, existing methods cannot make effective use of the background context for harmonization. For example, when the foreground object occupies a large part of the image, their performance commonly degrade.
    \item \textbf{Limited harmonization variability.} Current methods only consider simple low-dimensional color transfer functions to generate training (and even testing) data. That does not generalize well to real-world scenarios with drastically complex appearance discrepancies.
\end{itemize}


To tackle these limitations, we propose a new \textbf{S}elf-\textbf{S}upervised image \textbf{H}armonization framework, dubbed \textbf{SSH}. Different from  previous approaches~\cite{tsai2017deep,cong2020dovenet,cun2020improving} which directly take the compositing image as the input (as shown in Fig.~\ref{fig:Difference} left), the proposed SSH method  attempts to reformulate the harmonization problem from a representation fusion perspective. The proposed framework separately extracts the ``content" and ``appearance" representation from foreground and background images, and then aggregates these representations to synthesize the harmonized output. 
Based on this form, we introduce a novel \textit{dual data augmentation} engine to generate various synthesized data that can be directly taken as \textit{[foreground, background, pseudo GT]} triplets to support self-supervised training (shown in Fig.~\ref{fig:Difference} right). Meanwhile, we propose to adopt the 3D lookup table (LUT) to replace the traditional color transfer augmentation, which generates diverse visual examples on-the-fly. To this end, the proposed approach has no requirement for any foreground mask during training, and also allows us to leverage the \emph{entire} background image to generate high-quality harmonization results.

Previous methods evaluate their performance synthesized data~\cite{cong2020dovenet} or real-world data without ground truth~\cite{tsai2017deep}). In view of this gap, we build a new real-world, high-quality benchmark of harmonized composite images, that are retouched by professional Photoshop users. 
This collected dataset contains 216 composite images whose foregrounds include both the human portraits and general objects, while their backgrounds cover diverse environments such as mountains, rivers, buildings, sky and more (details are described in Sec.~\ref{sec:dataset}).  
Experiments demonstrate that SSH significantly outperforms state-of-the-art harmonization methods on the realistic data.

We summarize our contributions below:
\begin{itemize}[leftmargin=*]
\vspace{-0.5em}
    \item We propose the first \textbf{self-supervised} harmonization framework that needs neither human-annotated mask nor professionally created images for training. 
    \vspace{-0.2em}
    \item We develop a novel \textit{dual data augmentation} scheme, empowered by leveraging more complex 3D LUTs, to simulate more diverse and realistic training data on-the-fly.\vspace{-0.2em} 
    \item We collect the first-of-its-kind real-world benchmark set, containing 216 high-quality composite images that are professionally curated, to evaluate state-of-the-art image harmonization methods. Our method also significantly outperform existing approaches.
    \vspace{-0.5em}
\end{itemize}


\section{Related Works}


\textbf{Image Harmonization:}
Traditional image harmonization methods mainly target at better adjusting the low-level appearance statistics, such as color distribution~\cite{pitie2005n,reinhard2001color} and multi-scale features~\cite{jia2006drag,perez2003poisson,tao2010error}.
Besides traditional approaches, several recent works try to adopt learning-based method on harmonization task to better understand the context information between foreground and background images. Zhu \textit{et al.}~\cite{zhu2015learning} propose to use a discriminative model that can distinguish between natural images and composite images. Tsai \textit{et al.} firstly adopt segmentation mask as semantic information to train an end-to-end deep learning method. Cun \textit{et al.}~\cite{cun2020improving} propose the channel-wise and spatial-wise attention mechanism that further improve the visual quality of harmonized results. The most recent approaches DoveNet~\cite{cong2020dovenet} treats the image harmonization as a domain adaptation problem and successfully use adversarial learning to achieve notable performance. However, these methods \cite{tsai2017deep,cun2020improving,cong2020dovenet} mainly rely on the annotated segmentation mask to synthesize training pairs. Different from those methods, our method tackles the issue of expensive human annotation thanks to the superiority of self-supervision.

\textbf{Self-Supervised Learning:}
Self-supervised learning has been popular in high-level tasks by learning under pretext tasks \cite{gidaris2018unsupervised,noroozi2016unsupervised} or contrasting augmenting views \cite{chen2020simple,he2020momentum}. For low-level tasks, self-supervision is often implemented by self-generating synthetic data pairs, such as blur kernel in deblurring \cite{nah2017deep,kupyn2019deblurgan}, or bicubic interpolation in super resolution \cite{dong2015image}. Recently, \cite{laine2019high} applies self-supervised learning on image denoising without accessing clean reference images. However, most of these approaches meet severe challenges when tested on real-world data \cite{cai2019toward,rim2020real,jiang2021enlightengan}. To the best of our knowledge, SSH is the first self-supervised learning method on the image harmonization task.

\textbf{Comparing Image Harmonization and Style Transfer:}
Style transfer can be traced back to the seminal work of Gatys
\textit{et al.}~\cite{gatys2016image}, followed by many improved methods aiming at improving either transfer efficiency~\cite{huang2017arbitrary,johnson2016perceptual} or scalability~\cite{zhang2018multi,yang2019controllable,Gao_2020_WACV}. Nevertheless, its powerful ability on abstracting the texture feature makes it in general unsuitable for harmonizing the realistic \textit{photography images}. A particularly relevant route to us is the photorealistic style transfer~\cite{luan2017deep,li2018closed,yoo2019photorealistic}, which adds a photorealism regularization term on standard style transfer, resulting in a visually pleasing and photorealistic output. Although their goal is very similar to ours, existing works in this vein either need semantic segmentation masks to indicate specific regions~\cite{luan2017deep}, or require two input images to share a similar layout~\cite{li2018closed,yoo2019photorealistic} (\eg, building-to-building). Different from all of them, SSH can be adopted on arbitrary photography images without needing the segmentation mask or assuming similar layouts. We compare with the state-of-the-art photorealistic style transfer method $WCT^{2}$~\cite{yoo2019photorealistic} in our experiments to better illustrate the differences.

\section{Method} \label{sec:motivation}
\begin{figure*}[!ht]
\centering   
\includegraphics[width=\linewidth]{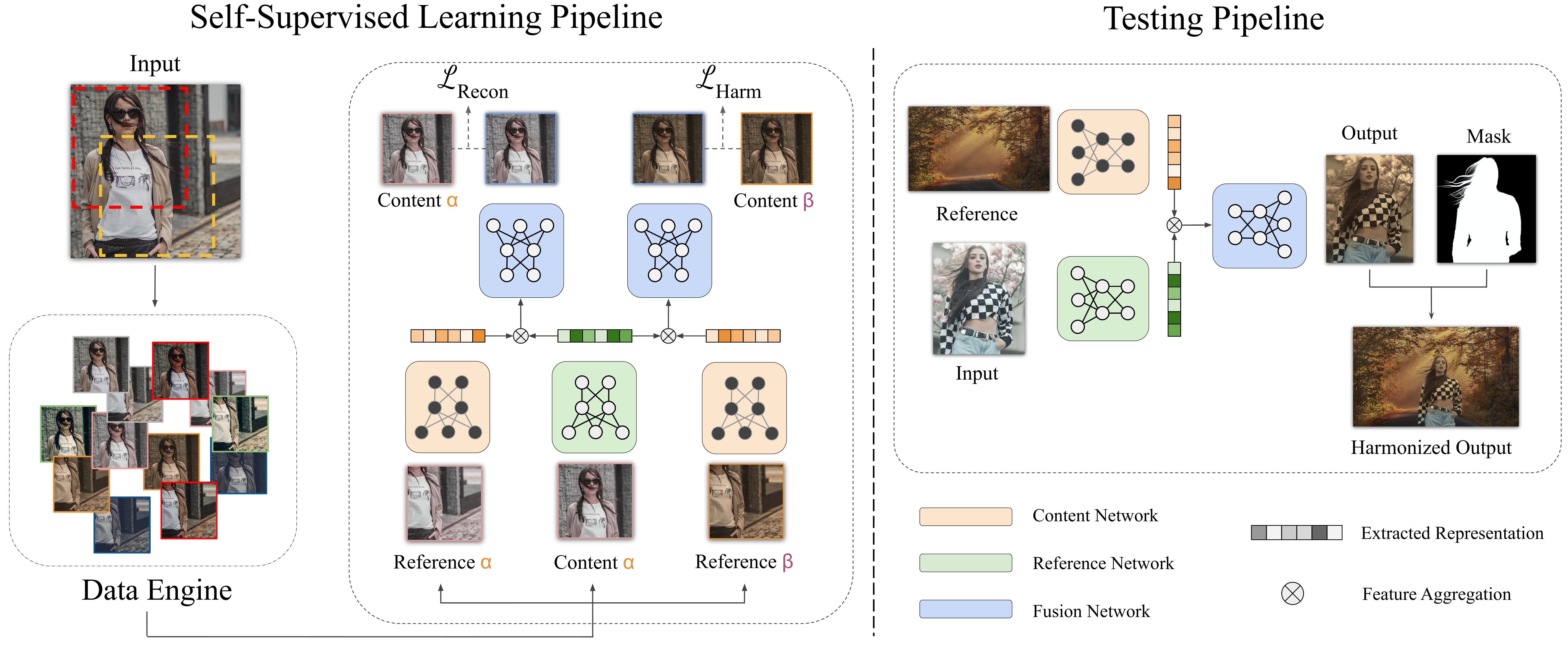}
\caption{\textbf{Details of Training and Testing Stage.} The left figure illustrate the main pipeline of our self-supervised framework. We firstly use the proposed dual data augmentation engine to generate \emph{Content} \textcolor{orange}{$\alpha$} \textcolor{purple}{$\beta$} and \emph{Reference} \textcolor{orange}{$\alpha$} \textcolor{purple}{$\beta$}, serving as the input of content network $G_C$ and reference network $G_S$ respectively (\textcolor{orange}{$\alpha$} and \textcolor{purple}{$\beta$} represents two different appearances after applying different 3D LUTs). After that the training pipeline learns to synthesize content \textcolor{purple}{$\beta$} from content \textcolor{orange}{$\alpha$} when the reference \textcolor{purple}{$\beta$} is given, and also learns to reconstruct content \textcolor{orange}{$\alpha$} itself when reference \textcolor{orange}{$\alpha$} is given. The translation and reconstruction process result in both the harmonization loss and the reconstruction loss. The right figure describe our testing stage. Noted that, \textbf{the human annotated mask is needed only in the testing stage for necessary composition.} } 
\label{fig:pipeline}
\end{figure*}

An overview of our proposed self-supervised harmonization framework (SSH) is shown in Fig.~\ref{fig:pipeline}.
Our main goal is to avoid the expensive human annotation that would be otherwise required for the harmonization task and define \textit{pseudo} ground truths that can instead serve as proxies for this task.
To this end, our method utilizes a content network and a reference network to extract representation about an input image's content and appearance respectively. The representations are extracted in different image crops. 
Then these features are concatenated and fed into a fusion network which aims at reconstructing the harmonized image. In addition,
processing the foreground and background images individually (rather than in a composite image \cite{cong2020dovenet}) allows us maximize information and avoid the background occlusion problem (Fig.~\ref{fig:back}).


We observe that the different crops from one image tend to share the same appearance (color, lighting condition, and contrast) since they are captured from the same environment (lighting, weather condition) and camera setting. Thus the different crops and their different appearance version with proper perturbations can serve as a pseudo triplet (the foreground, background and ground truth) for the content and reference networks.
Inspired by this observation, we propose a dual data augmentation scheme, consisting of a content augmentation and an appearance augmentation. 
We will first introduce the dual data augmentation strategy, then proceed to explaining the details of our self-supervised framework and the data collection procedure.

\subsection{Dual Data Augmentation}
The goal of dual data augmentation is to provide pseudo training triplets that contain various appearance and also mimic real testing scenarios.
In each iteration, we perform both content and appearance augmentations. The content augmentation samples two different crops with some overlapping region. 
Meanwhile, the appearance augmentation applies multiple 3D color lookup table (LUT) for the given image to obtain its corresponding stylized images. 
\subsubsection{Content Augmentation}
To simulate real testing scenarios where the foreground and background images are totally different, we apply content augmentation in the data synthesis process. The content augmentation adopts a simple yet effective multi-cropping method to generate different crops of one original image. The cropping size ranges from a local region to a global region, and thus can mimic diverse environments which reduces the gap between synthetic data and real testing data. The bottom left image in Fig.~\ref{fig:pipeline} shows a typical example where we can obtain different looks from an image by this cropping method. 

\subsubsection{Appearance Augmentation} \label{sec:LUT}
Synthesizing data with various appearance is a common step in training image harmonization models \cite{cong2020dovenet,tsai2017deep,cun2020improving}. Existing approaches either choose to adopt a single color transfer method \cite{tsai2017deep}, or simply extend it to several different color transfer approaches \cite{cong2020dovenet}. However, in real harmonization scenarios, the appearance mismatch between the foreground and background images can be significantly more complex and include contrast, brightness, and saturation differences. To tackle this issue, we propose to use a 3D color lookup table as the basic transformation approach.

\textbf{A 3D Color Lookup Table (LUT)} maps one color space to another, and is widely used in film post-production industry. It is essentially a 3D-to-3D mapping that can transform any RGB color of an input image to any other RGB color. 
It can also represents functions like contrast enhancement where the tonal range of the input image is manipulated. 

A LUT has a number of advantages including: 1) Unlike simple color transfer functions, it can represent complex appearance adjustments; for example, the bottom left image in Fig.~\ref{fig:pipeline} shows that the LUT can provide a non-linear transformation where the appearance of different parts of the image are transformed differently (e.g., the skin can be transformed to red color while the T-shirt remains white color), 2) Given one input image, there exist hundreds of LUTs that can be applied to generate its stylized versions and dramatically enrich the training data, and 3) LUT processing is real-time so that it can be applied as an on-the-fly data augmentation strategy.

\subsection{Self-Supervised Framework} 
The proposed self-supervised framework SSH takes the foreground as content image $C$ and the background as reference image $R$. We adopt a reference network $G_r$ to capture the appearance representation (color, brightness, and contrast etc.) from reference image $R$, and a content network $G_c$ to capture the content representation (structure, texture etc.) from content input $C$. Then the fusion network $F$ aggregates the appearance and content representation and learns to synthesize the output $C^{'}$, so that the output $C^{'}$ matches the appearance of reference image $R$ and preserves the content of content image $C$. We formulate the process as following:
\begin{equation} \label{eq:process}
    C^{'} = F(G_c(C), G_r(R))
\end{equation}

Using the proposed dual data augmentation, we can generate pseudo triplets [foreground, background, and ground truth] from one image, and adopt it for training SSH. As shown in the left part of Fig.~\ref{fig:pipeline}, we generate two image with their appearance perturbed by two different 3D LUTs separately (denoted by \textcolor{orange}{$\alpha$} and \textcolor{purple}{$\beta$}).
Therefore, content crops (denoted as $C_{\textcolor{orange}{\alpha}}$, $C_{\textcolor{purple}{\beta}}$) and reference crops (denoted as $R_{\textcolor{orange}{\alpha}}$, $R_{\textcolor{purple}{\beta}}$) corresponding to them can be obtained. 

Here $C_{\textcolor{orange}{\alpha}}$ contain similar appearance information with $R_{\alpha}$ and the same content information as $C_{\textcolor{purple}{\beta}}$. Then the network is expect to map $C_{\textcolor{orange}{\alpha}}$ to $C_{\textcolor{purple}{\beta}}$ when taking $R_{\textcolor{purple}{\beta}}$ as the reference, and reconstruct the $C_{\textcolor{orange}{\alpha}}$ to itself when taking $R_{\textcolor{orange}{\alpha}}$ as the reference. This special design simulate real testing scenarios where the foreground and background does not share the same content while the output is expected to have same the appearance as the background and the same content as the foreground. The mapping process and reconstruction process use the following harmonization loss $L_{harm}$ and reconstruction loss $L_{harm}$:

\begin{equation}
    C_{\textcolor{purple}{\beta}}^{'} = F(G_c(C_{\textcolor{orange}{\alpha}}), G_s(R_{\textcolor{purple}{\beta}}))
\end{equation}
\begin{equation}
    C_{\textcolor{orange}{\alpha}}^{'} = F(G_c(C_{\textcolor{orange}{\alpha}}), G_s(R_{\textcolor{orange}{\alpha}}))
\end{equation}
\begin{equation} \label{eq: loss_1}
    {L_{harm}} = {||C_{\textcolor{purple}{\beta}}^{'} - C_{\textcolor{purple}{\beta}}||}^2
\end{equation}
\begin{equation} \label{eq: loss_1}
    {L_{recon}} = {||C_{\textcolor{orange}{\alpha}}^{'} - C_{\textcolor{orange}{\alpha}}||}^2
\end{equation} 

\begin{figure}[t]
\centering   
\includegraphics[width=\linewidth]{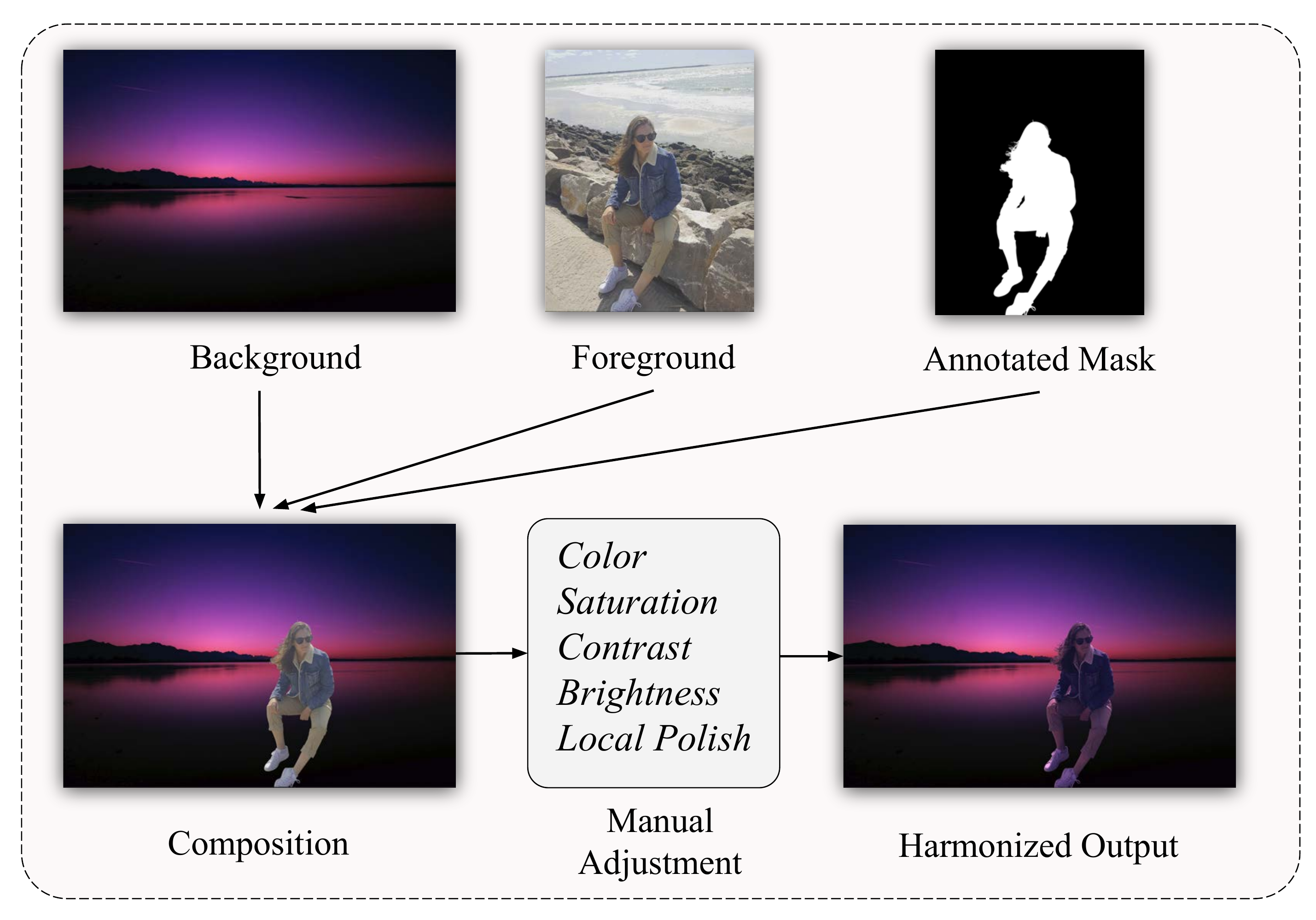}
\caption{\textbf{The Main Pipeline of Dataset Collection.} The detailed pipeline of how the composite image is retouched by professional user is shown above. It includes both the object annotation and appearance matching (brightness, color/saturation, contrast adjustment, and local polish).}
\label{fig:adjustment}
\end{figure}

Since the reference network $G_r$ needs to capture the appearance information from given image, it is expected to extract similar representation when receiving different crops with the same appearance. Meanwhile, the content network $G_c$ is expected to capture the same feature given the same crops with different appearance. Thus we design another disentanglement loss $L_{dis}$ formulated as following:
\begin{equation} \label{eq: dis_loss}
    {L_{dis}} = {||G_c(C_{\textcolor{orange}{\alpha}}) - G_c(C_{\textcolor{purple}{\beta}})||}^2 + {||G_s(C_{\textcolor{orange}{\alpha}}) - G_s(R_{\textcolor{orange}{\alpha} })||}^2
\end{equation}
Combining with harmonization loss and reconstruction loss defined above, the overall loss function for training SSH is thus written as:
\begin{equation} \label{eq: loss_1}
    {Loss} = L_{harm} + w_1*L_{recon}+ w_2*L_{dis}
\end{equation}
Where $w_1$ and $w_2$ is set to 0.4 and 0.05 in experiments. Therefore, the proposed SSH can translate a image to a synthesized one that the synthesized output can match the appearance of another images. As shown in the right part of Fig.~\ref{fig:pipeline}, only in the testing stage, we adopt the object mask to composite synthesized output and background/reference image to generate the harmonized output


\begin{figure*}[ht]
  \centering
\begin{tabular}[!t]{ccccccc}
  \includegraphics[trim={0 0 0 0},clip,width=0.133\linewidth]{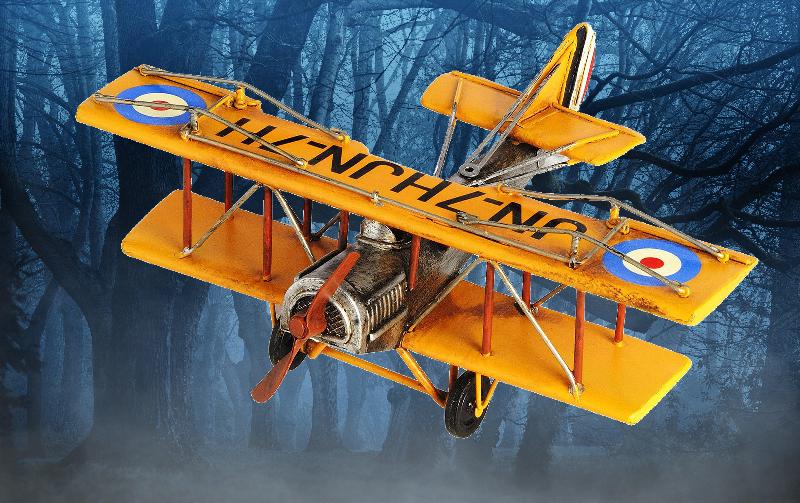} \hspace{-1.2em} &  
  \includegraphics[trim={0 0 0 0},clip,width=0.133\linewidth]{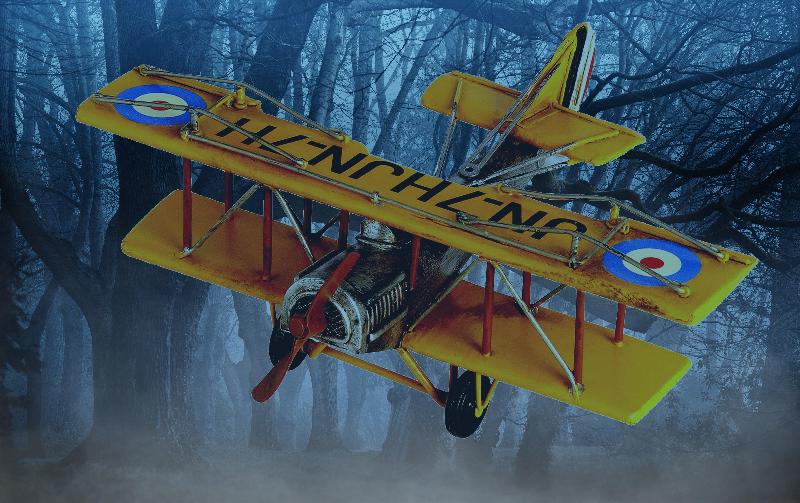} \hspace{-1.2em} & 
  \includegraphics[trim={0 0 0 0},clip,width=0.133\linewidth]{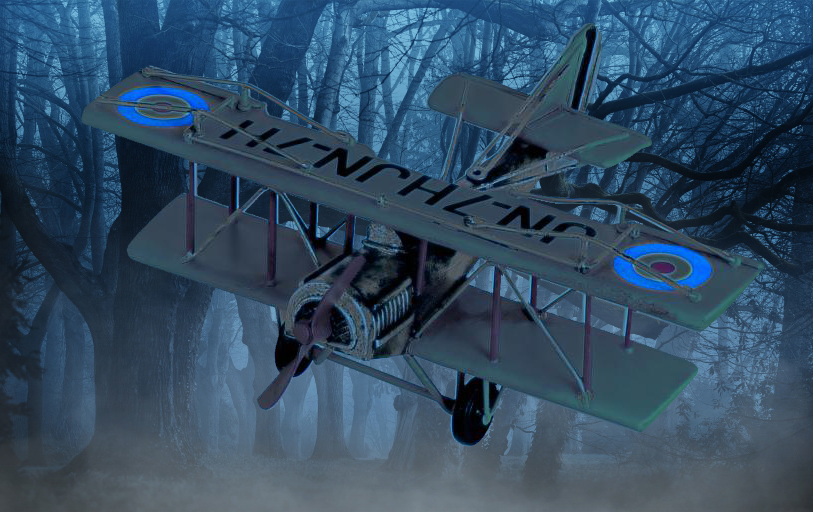} \hspace{-1.2em} &  
  \includegraphics[trim={0 0 0 0},clip,width=0.133\linewidth]{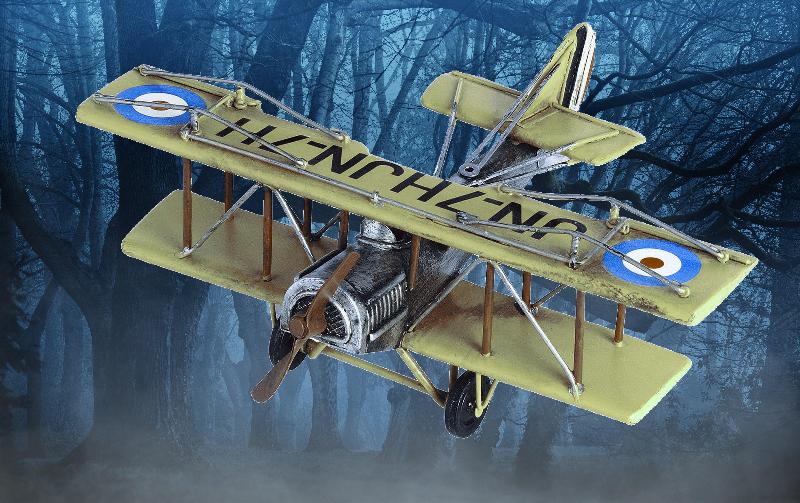}
   \hspace{-1.2em} & 
  \includegraphics[trim={0 0 0 0},clip,width=0.133\linewidth]{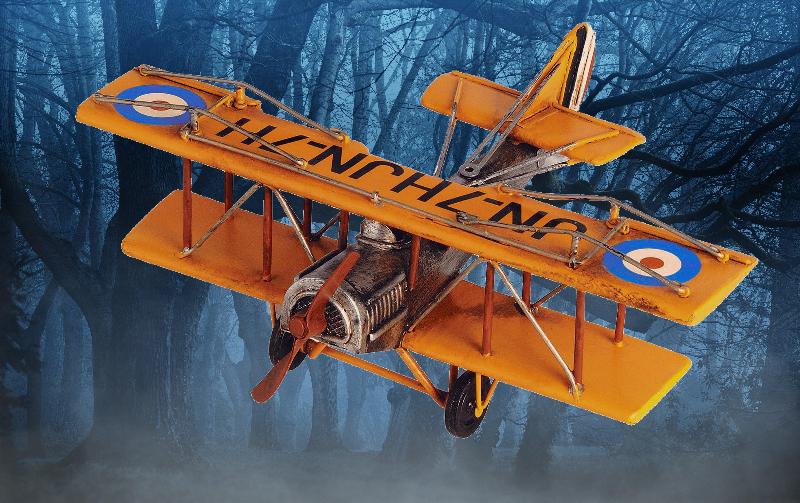} \hspace{-1.2em} & 
  \includegraphics[trim={0 0 0 0},clip,width=0.133\linewidth]{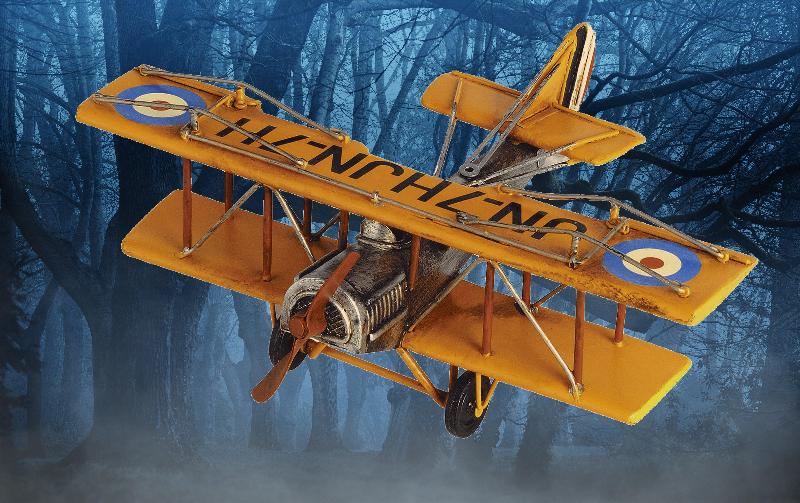} \hspace{-1.2em} &
  \includegraphics[trim={0 0 0 0},clip,width=0.133\linewidth]{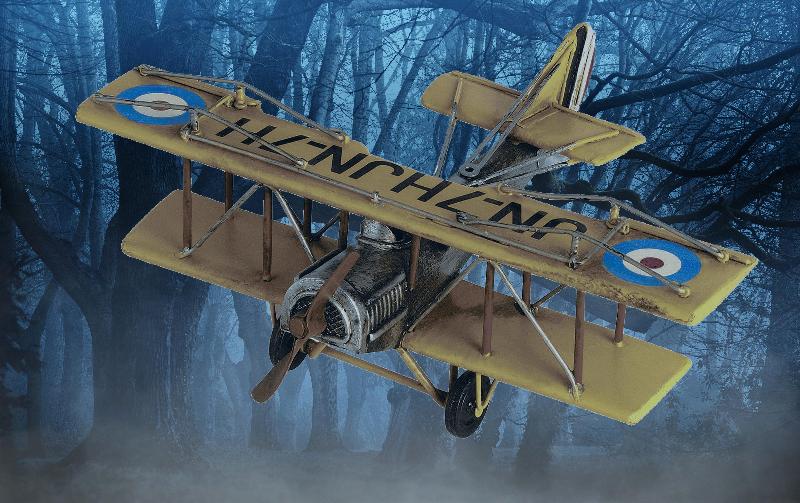}
  \\
  \includegraphics[trim={0 0 10cm 0},clip,width=0.133\linewidth]{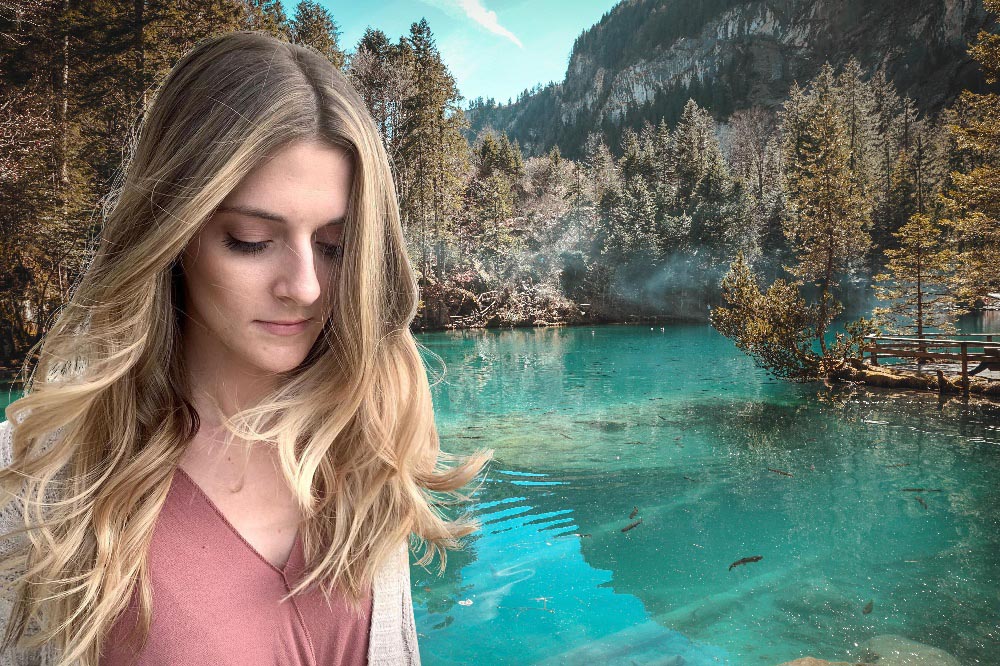} \hspace{-1.2em} &  
  \includegraphics[trim={0 0 10cm 0},clip,width=0.133\linewidth]{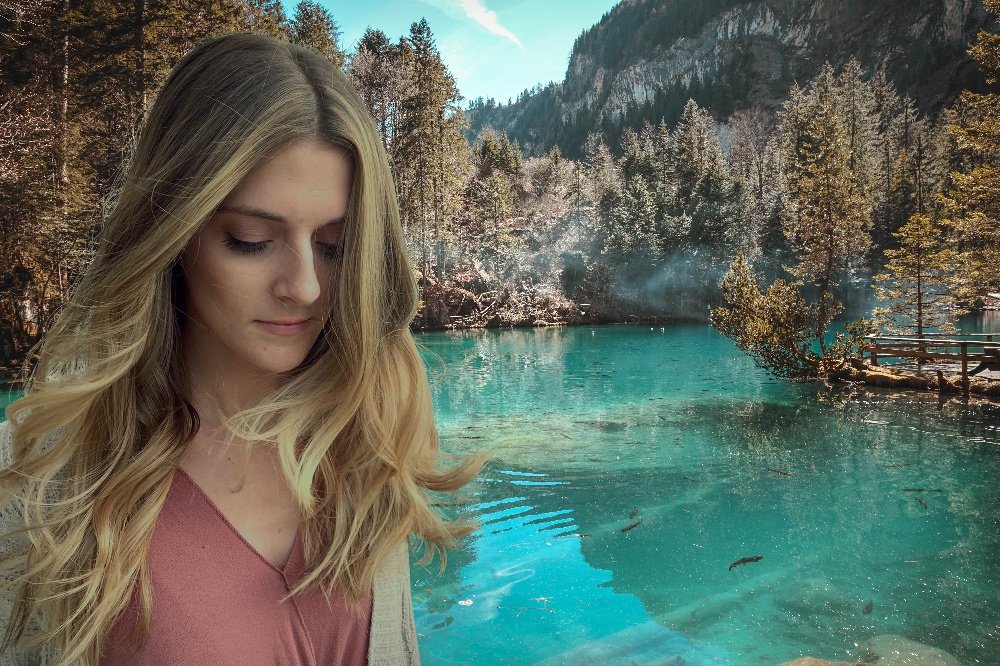} \hspace{-1.2em} & 
  \includegraphics[trim={0 0 7.6cm 0},clip,width=0.133\linewidth]{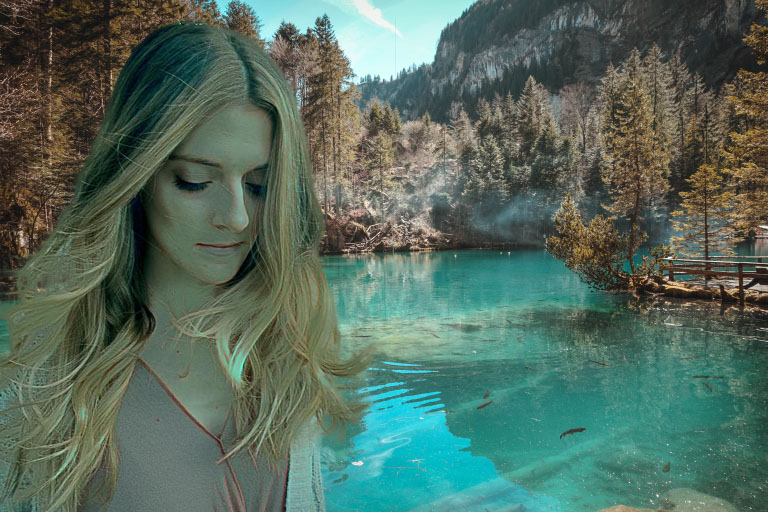} \hspace{-1.2em} &  
  \includegraphics[trim={0 0 10cm 0},clip,width=0.133\linewidth]{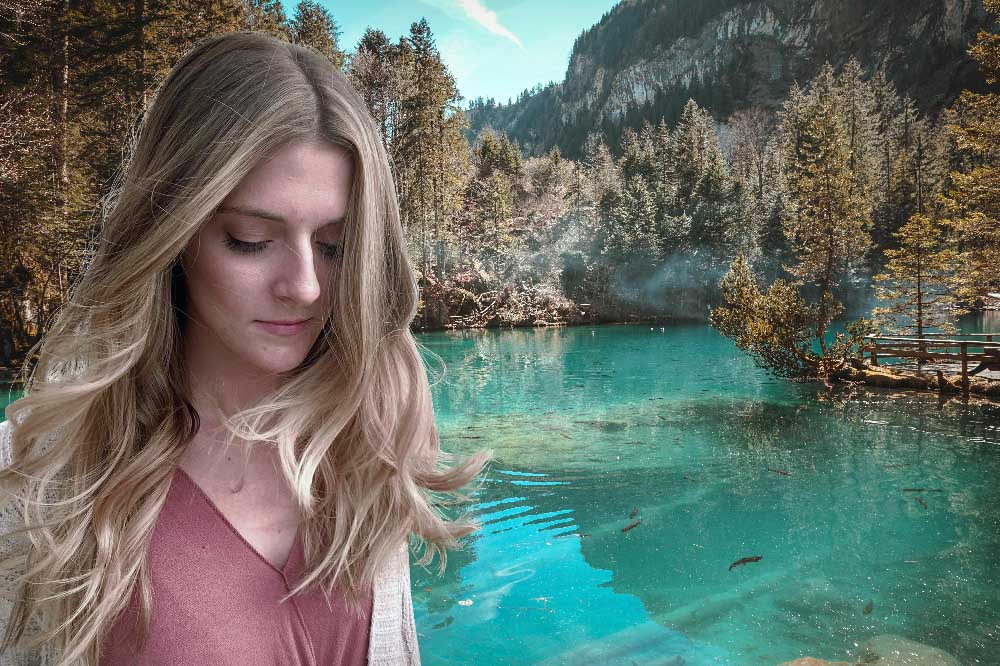} \hspace{-1.2em} &  
  \includegraphics[trim={0 0 10cm 0},clip,width=0.133\linewidth]{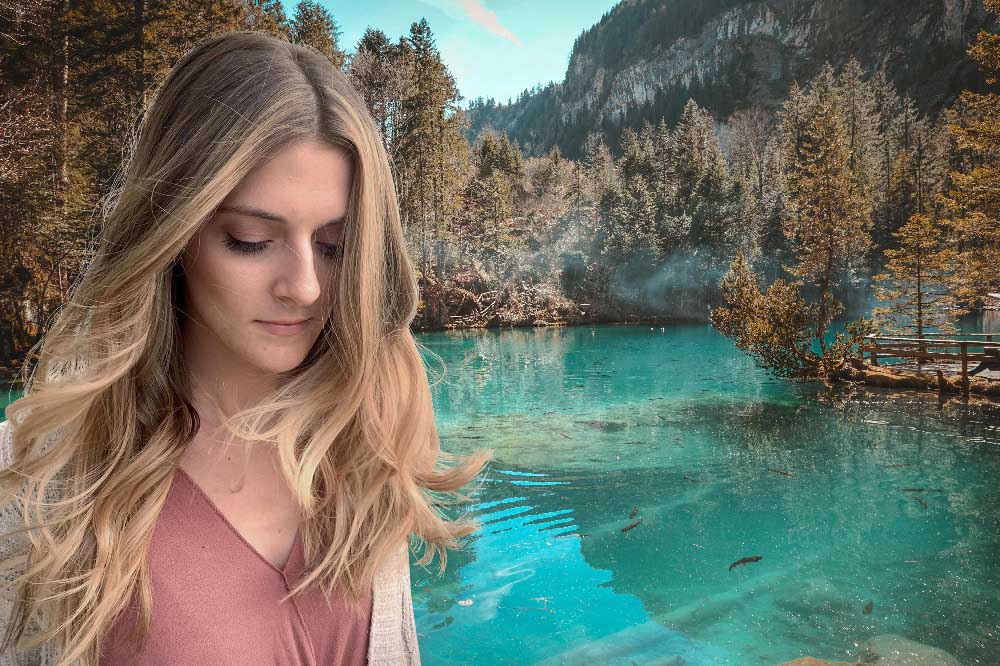} \hspace{-1.2em} & 
  \includegraphics[trim={0 0 10cm 0},clip,width=0.133\linewidth]{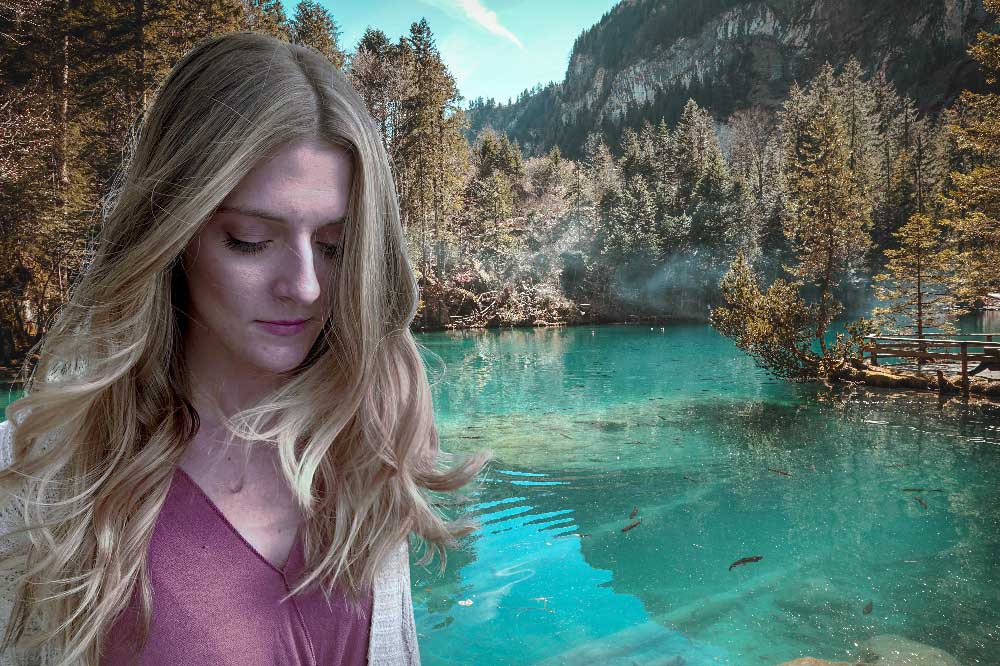} \hspace{-1.2em} &
  \includegraphics[trim={0 0 10cm 0},clip,width=0.133\linewidth]{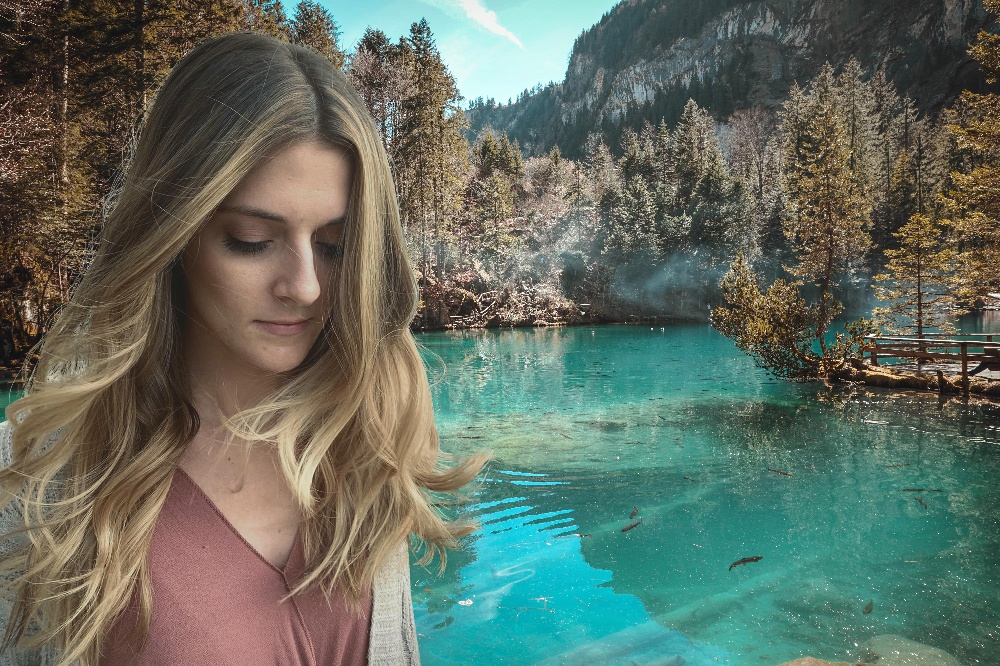}
  \\
  \includegraphics[trim={5cm 0 5cm 5cm 5cm},clip,width=0.133\linewidth]{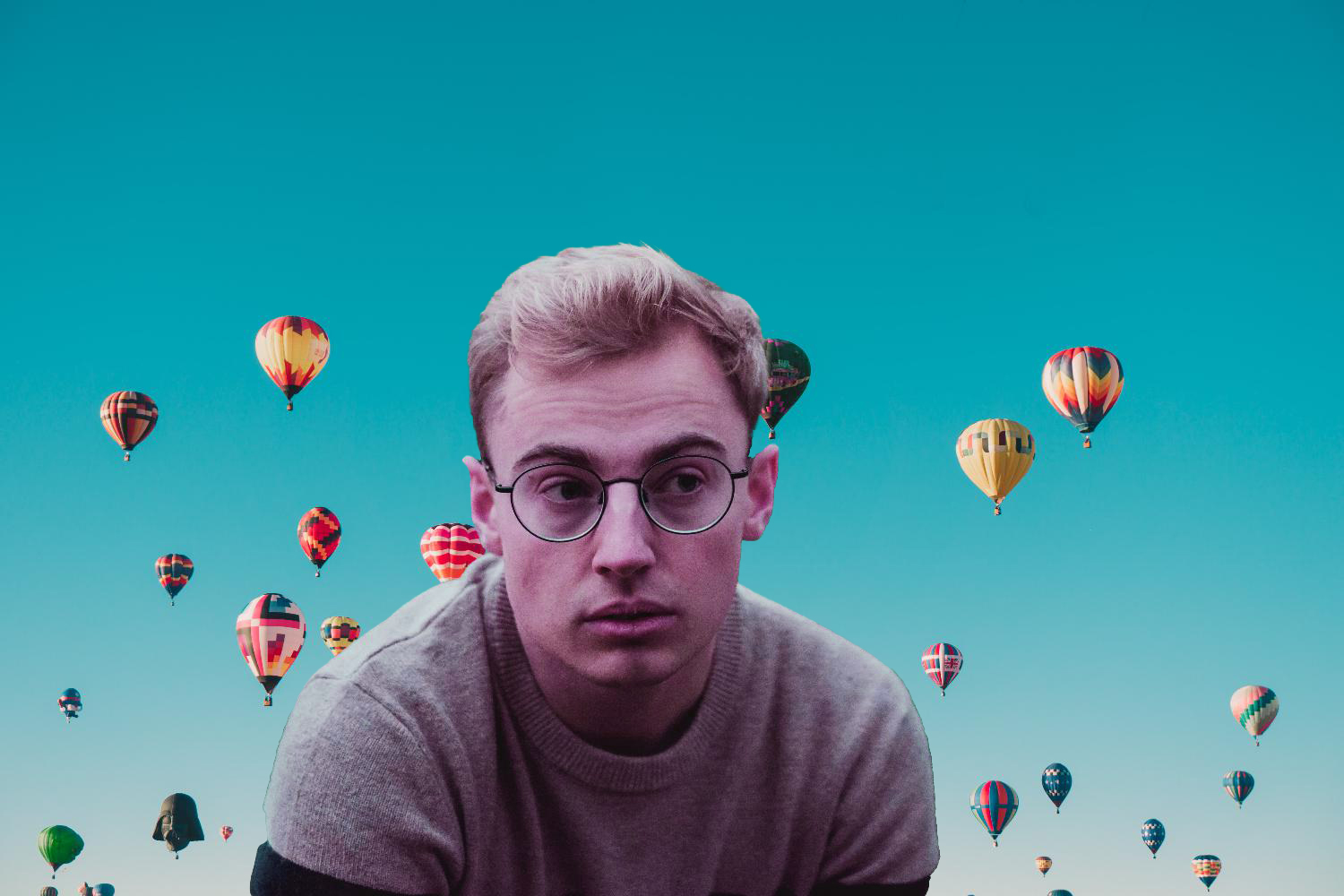} \hspace{-1.2em} &  
  \includegraphics[trim={5cm 0 5cm 5cm 5cm},clip,width=0.133\linewidth]{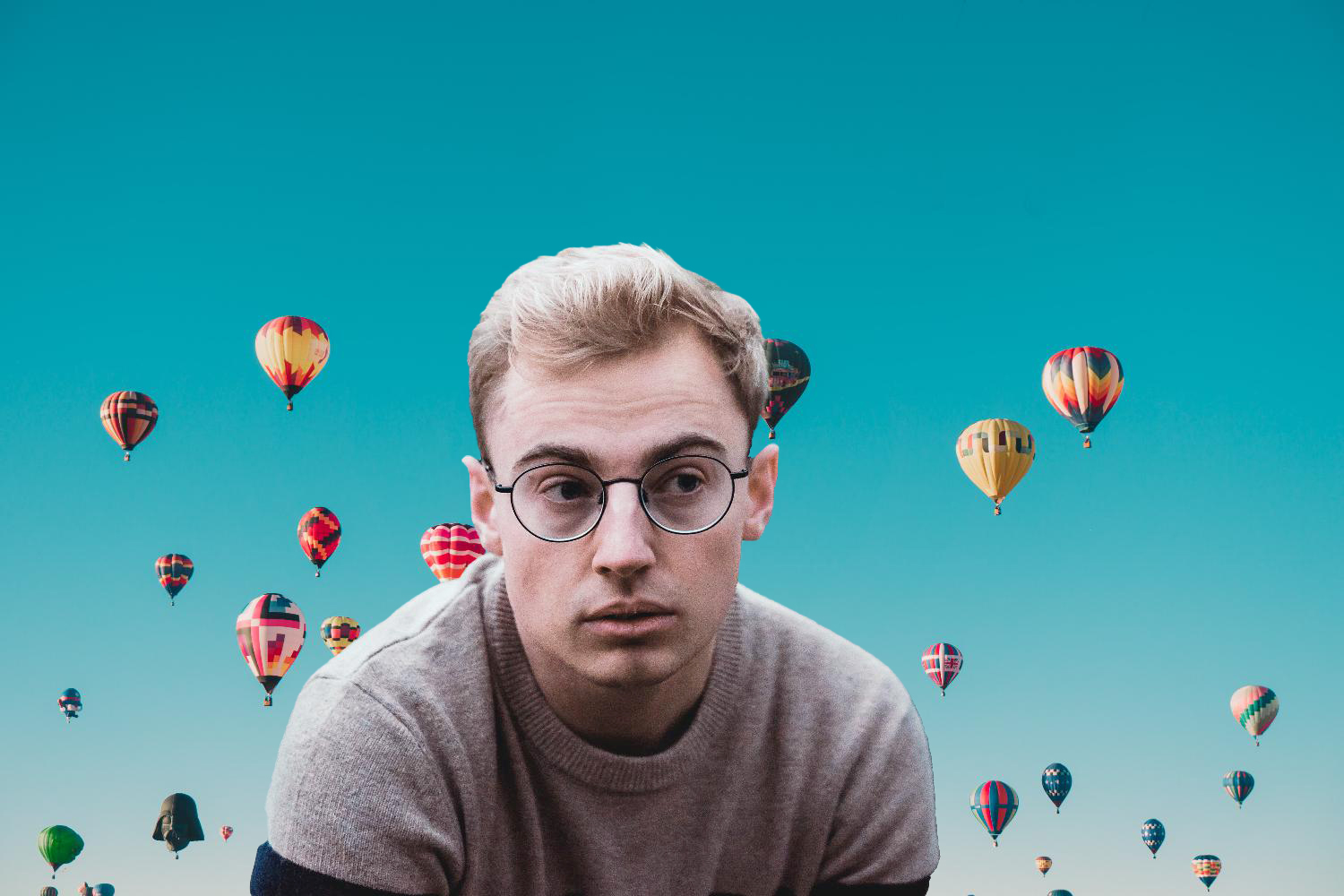} \hspace{-1.2em} &  
  \includegraphics[trim={2.5cm 0 2.5cm 2.5cm 2.5cm},clip,width=0.133\linewidth]{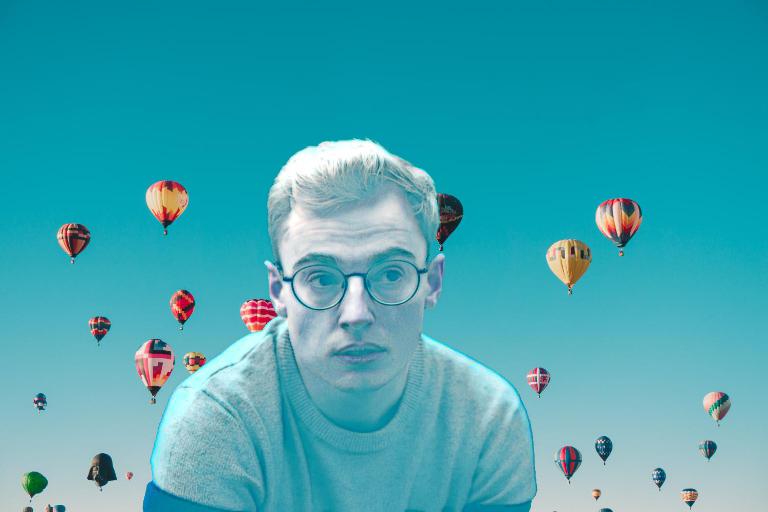} \hspace{-1.2em} &  
  \includegraphics[trim={5cm 0 5cm 5cm 5cm},clip,width=0.133\linewidth]{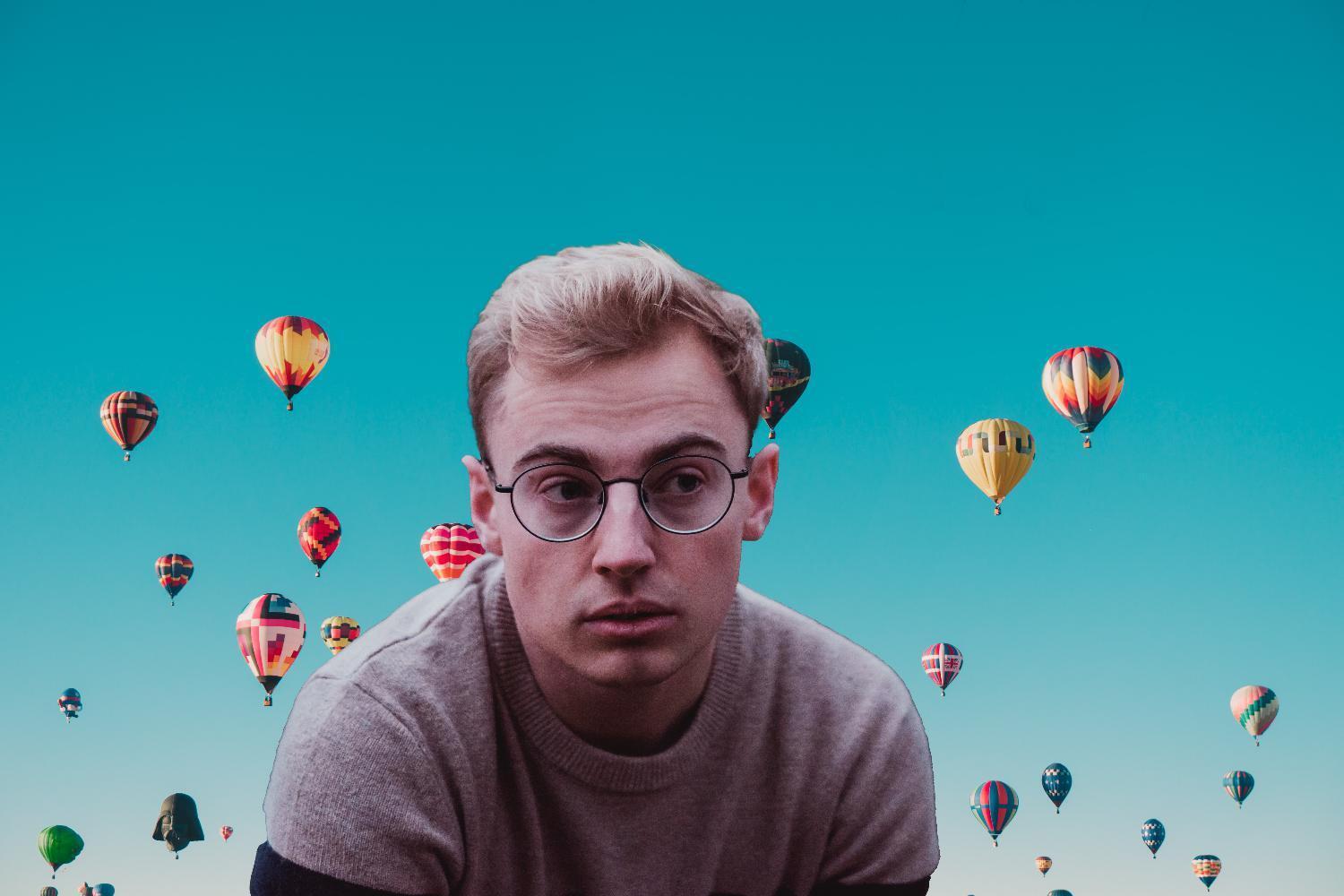} \hspace{-1.2em} &  
  \includegraphics[trim={5cm 0 5cm 5cm 5cm},clip,width=0.133\linewidth]{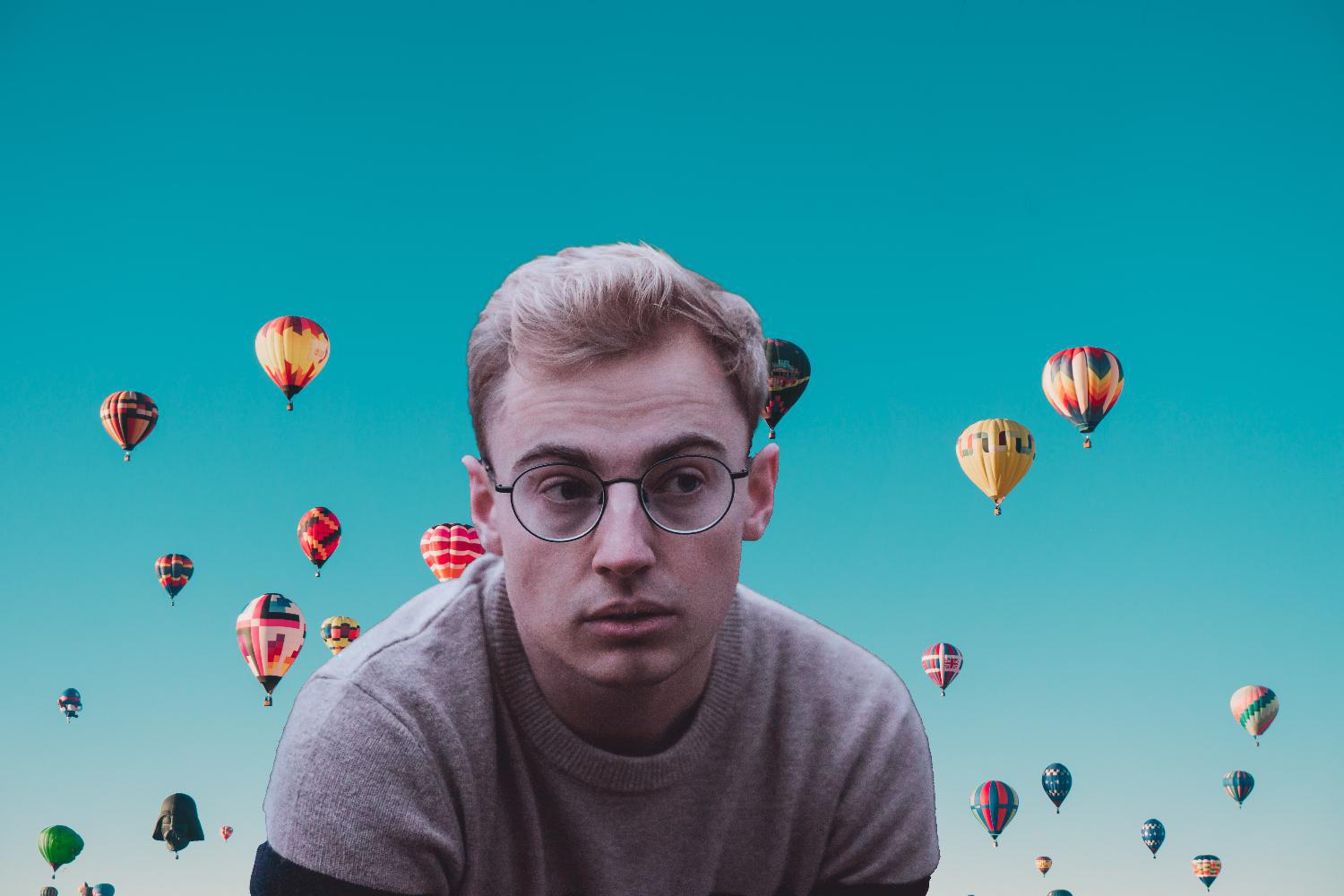} \hspace{-1.2em} &  
  \includegraphics[trim={5cm 0 5cm 5cm 5cm},clip,width=0.133\linewidth]{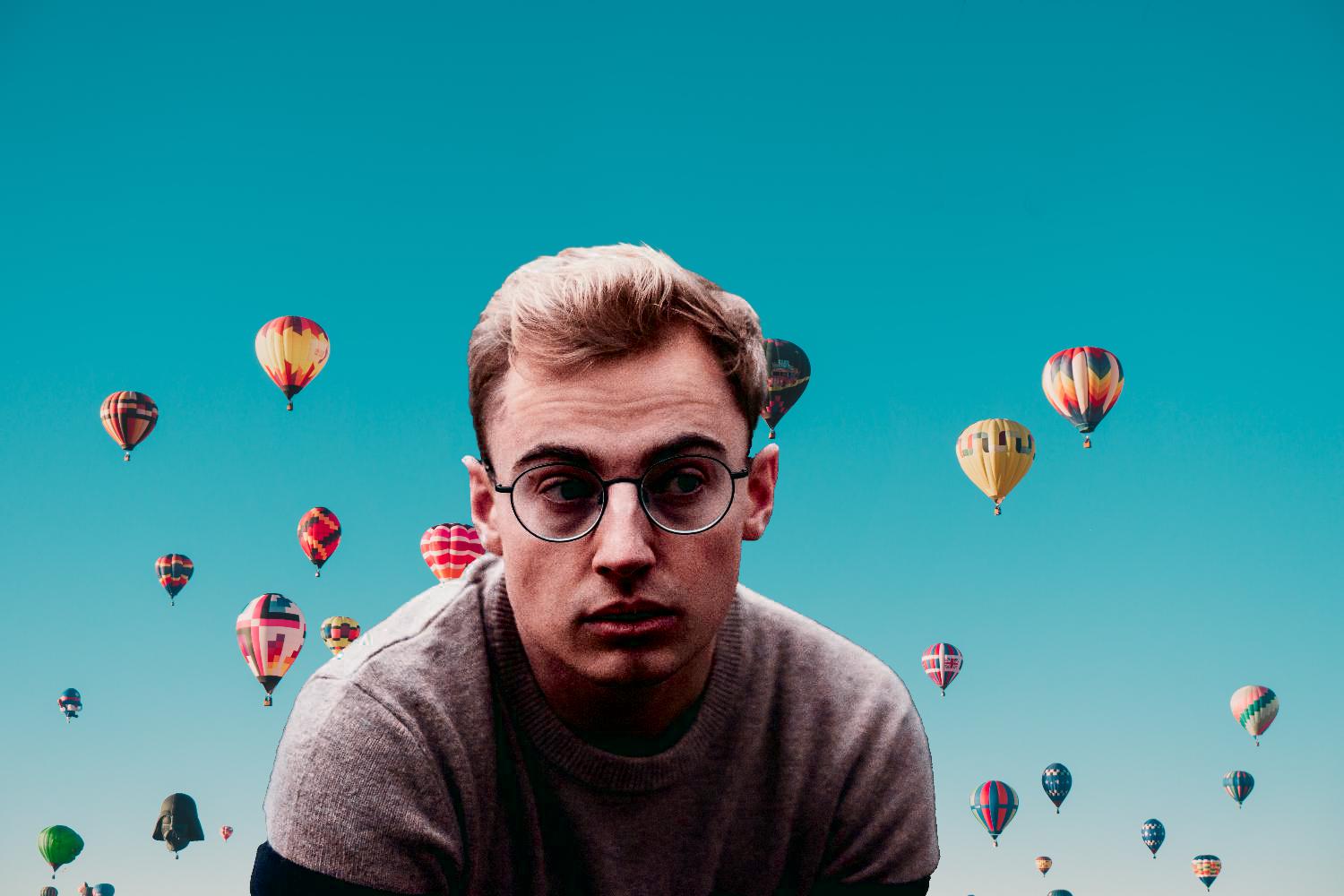} \hspace{-1.2em} &  
  \includegraphics[trim={5cm 0 5cm 5cm 5cm},clip,width=0.133\linewidth]{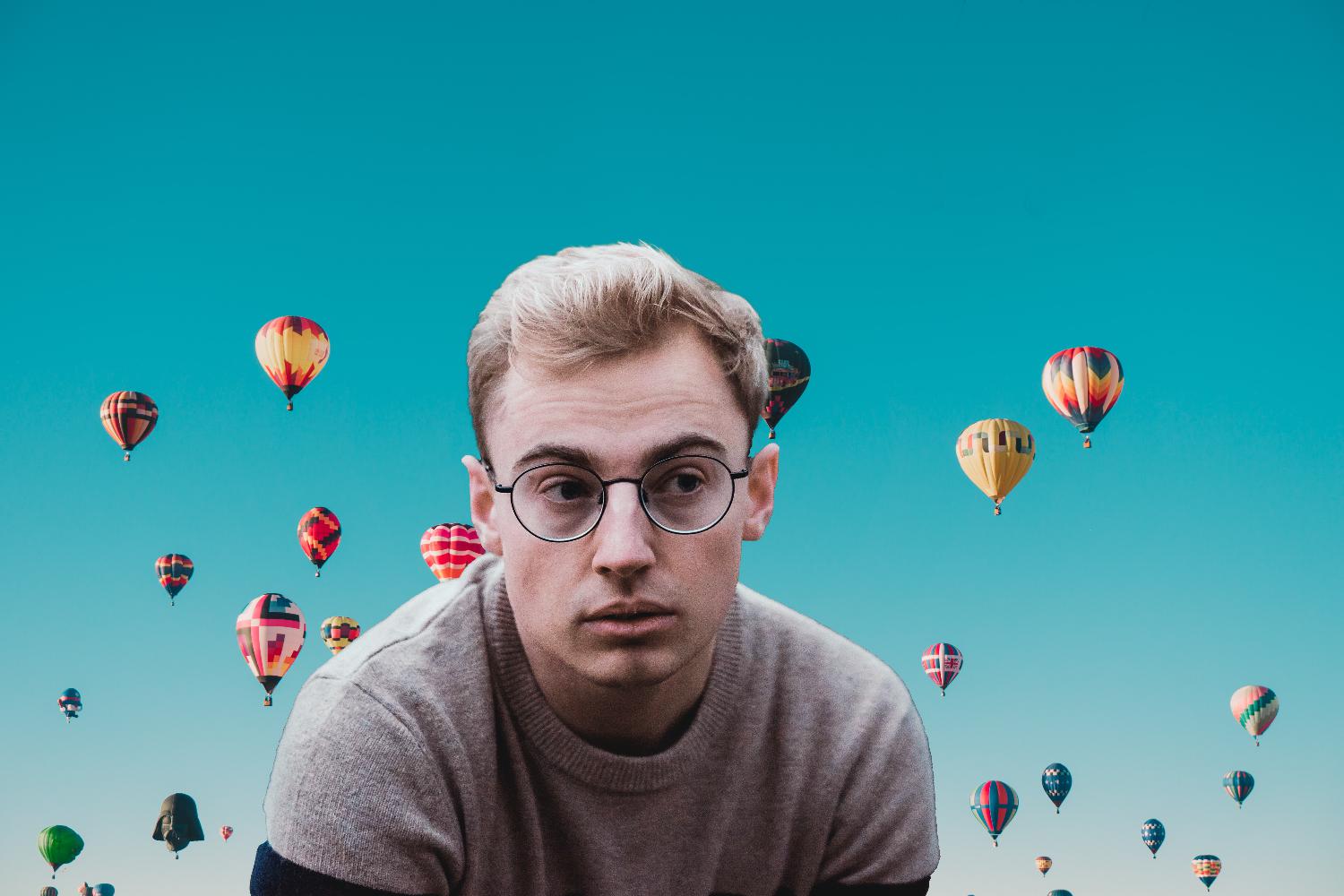}  

  \\
  
  \includegraphics[trim={0cm 10cm 0cm 0cm},clip,width=0.133\linewidth]{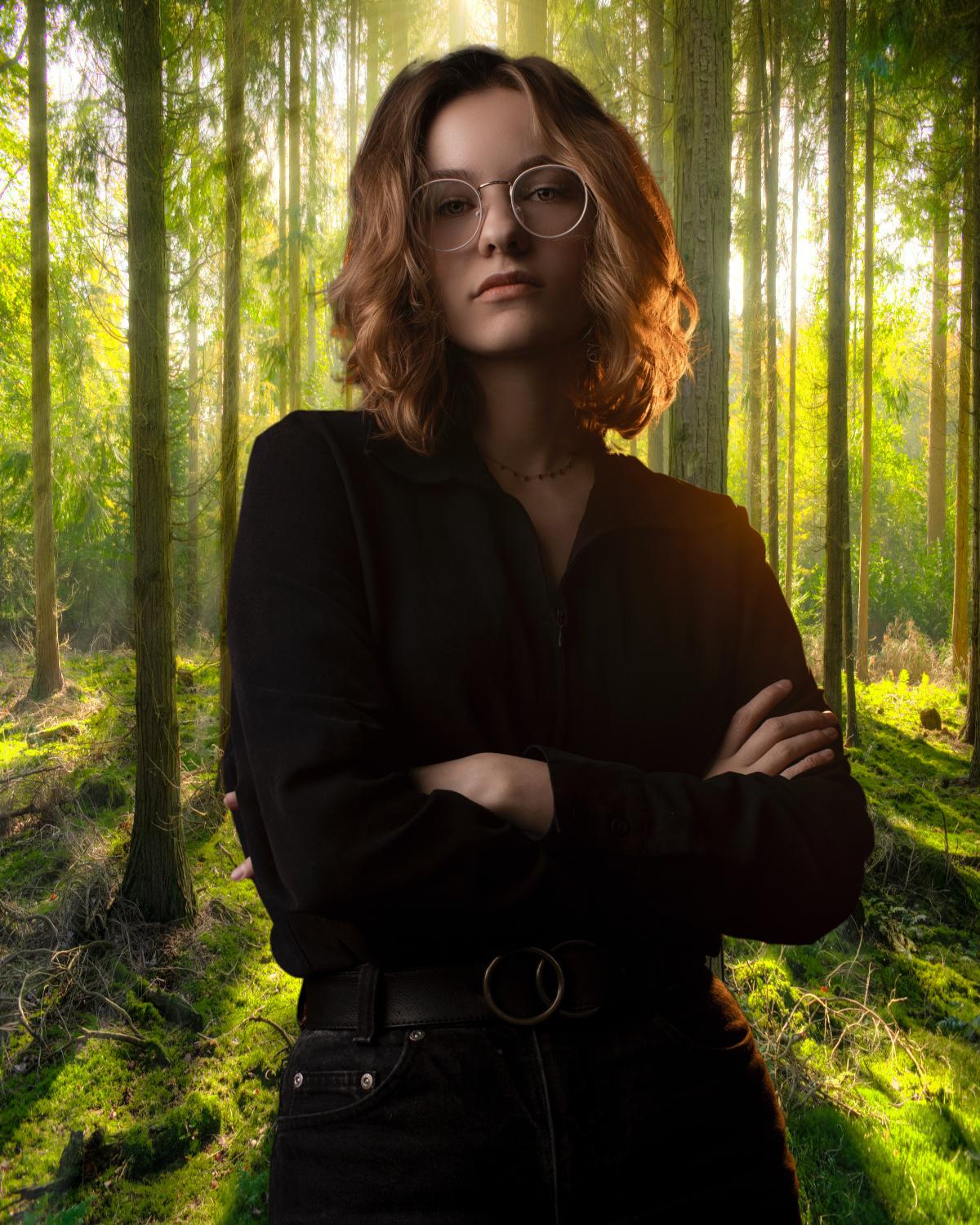} \hspace{-1.2em} &  
  \includegraphics[trim={0cm 10cm 0cm 0cm},clip,width=0.133\linewidth]{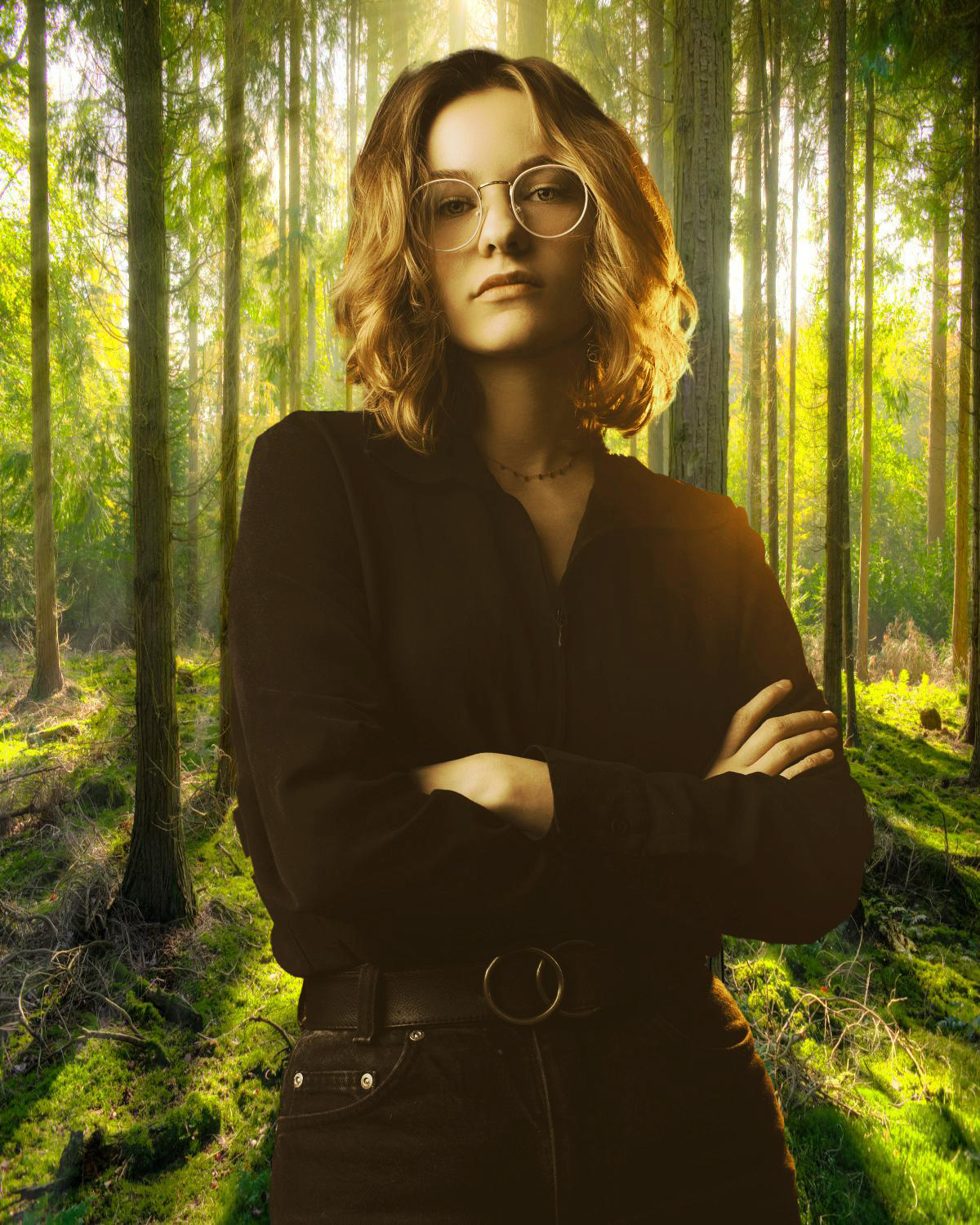} \hspace{-1.2em} &  
  \includegraphics[trim={0cm 4.2cm 0cm 0cm},clip,width=0.133\linewidth]{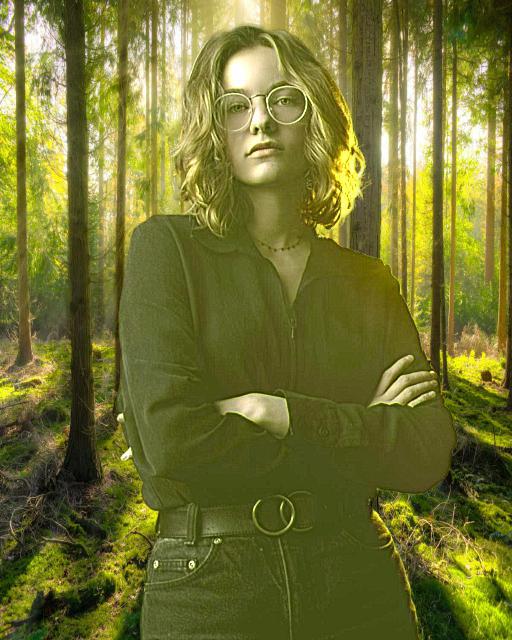} \hspace{-1.2em} &  
  \includegraphics[trim={0cm 10cm 0cm 0cm},clip,width=0.133\linewidth]{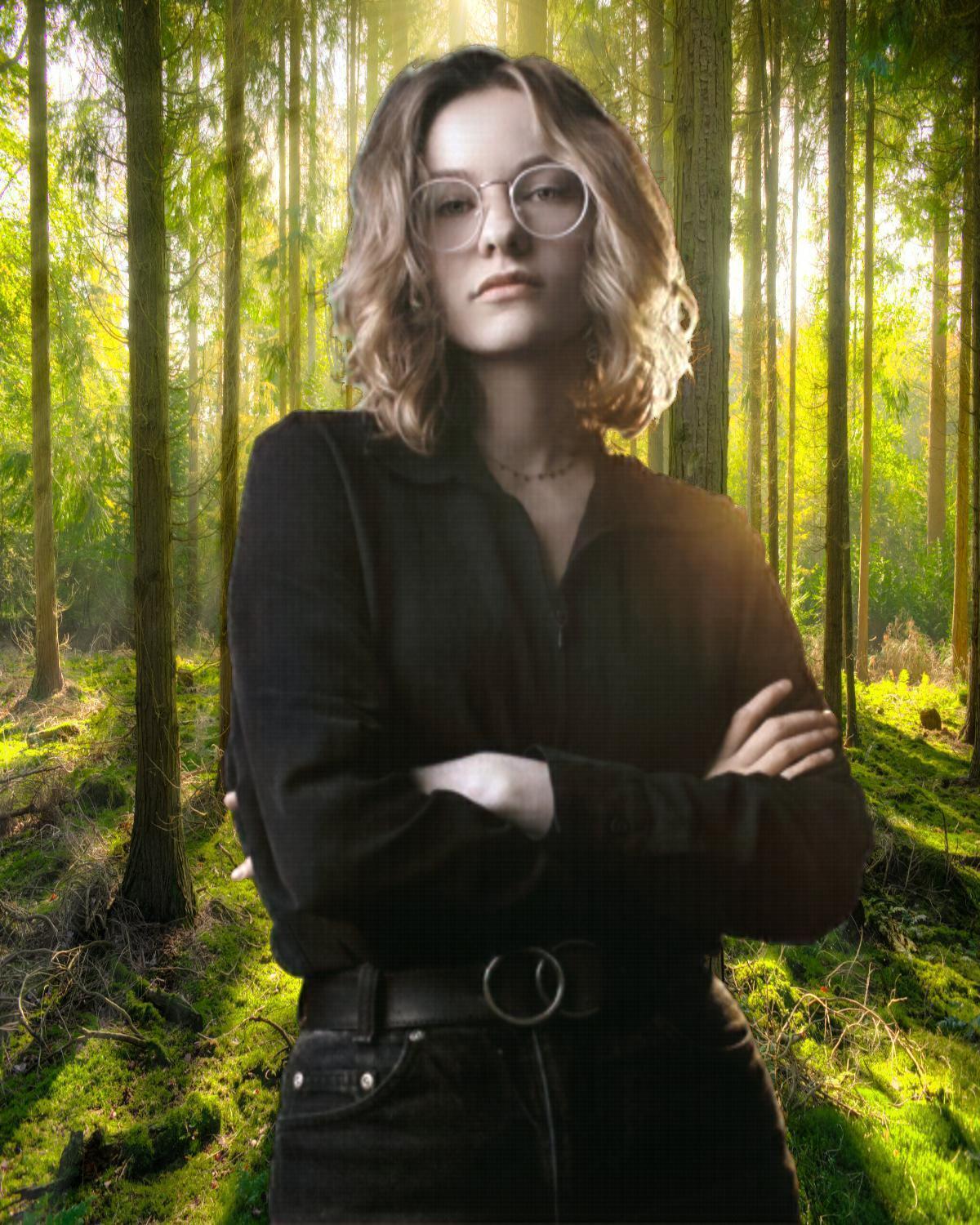} \hspace{-1.2em} &  
  \includegraphics[trim={0cm 10cm 0cm 0cm},clip,width=0.133\linewidth]{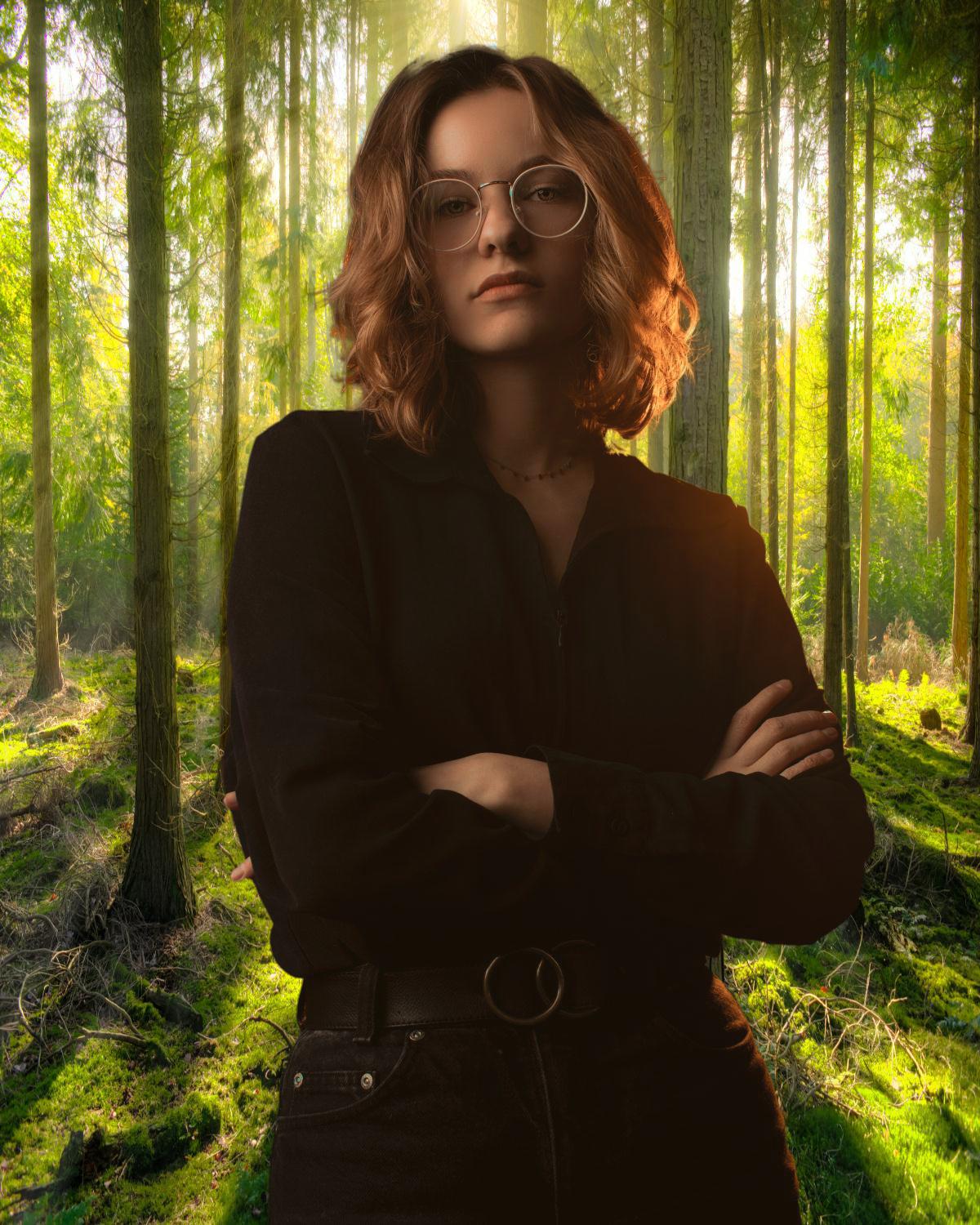} \hspace{-1.2em} &  
  \includegraphics[trim={0cm 10cm 0cm 0cm},clip,width=0.133\linewidth]{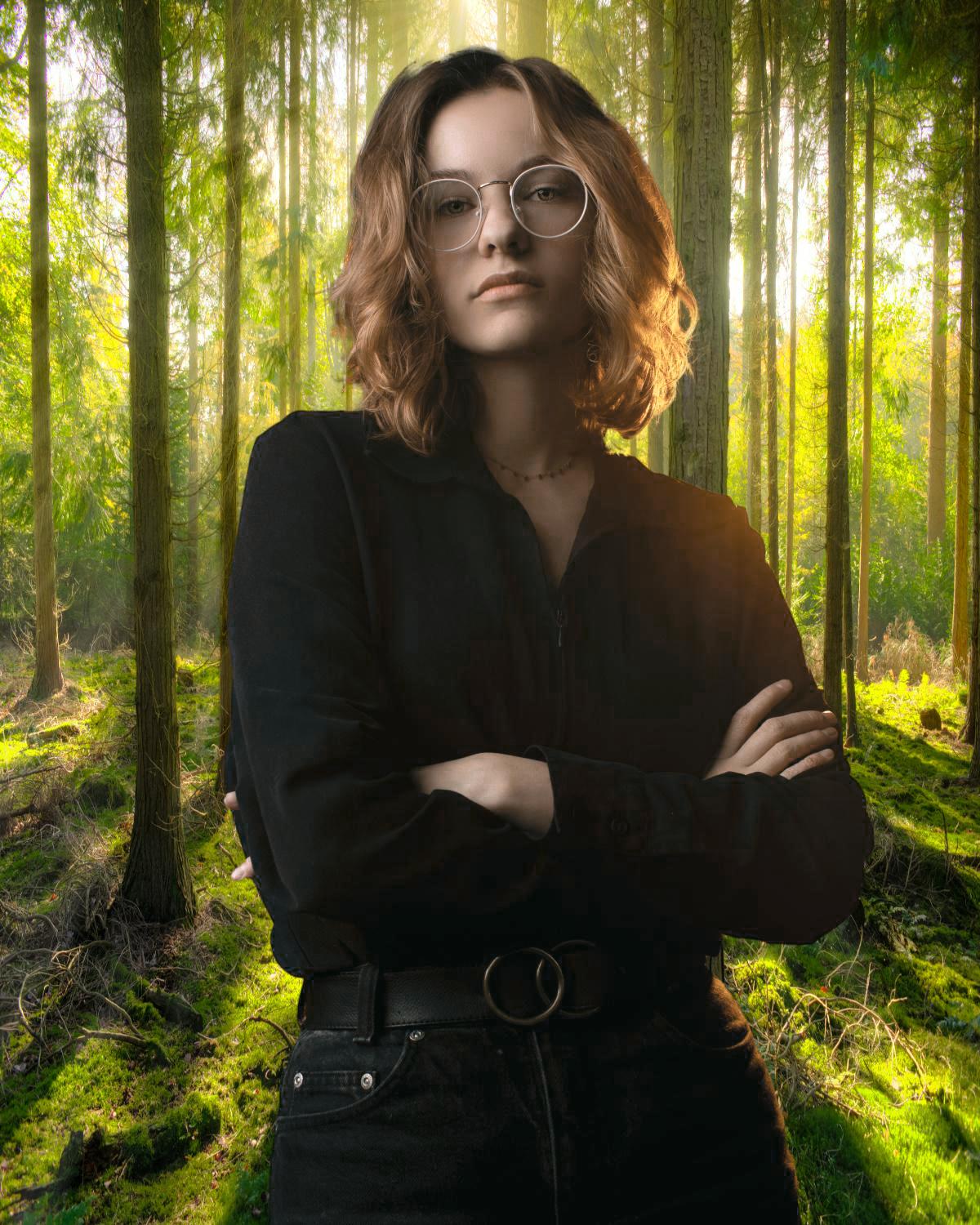} \hspace{-1.2em} &  
  \includegraphics[trim={0cm 10cm 0cm 0cm},clip,width=0.133\linewidth]{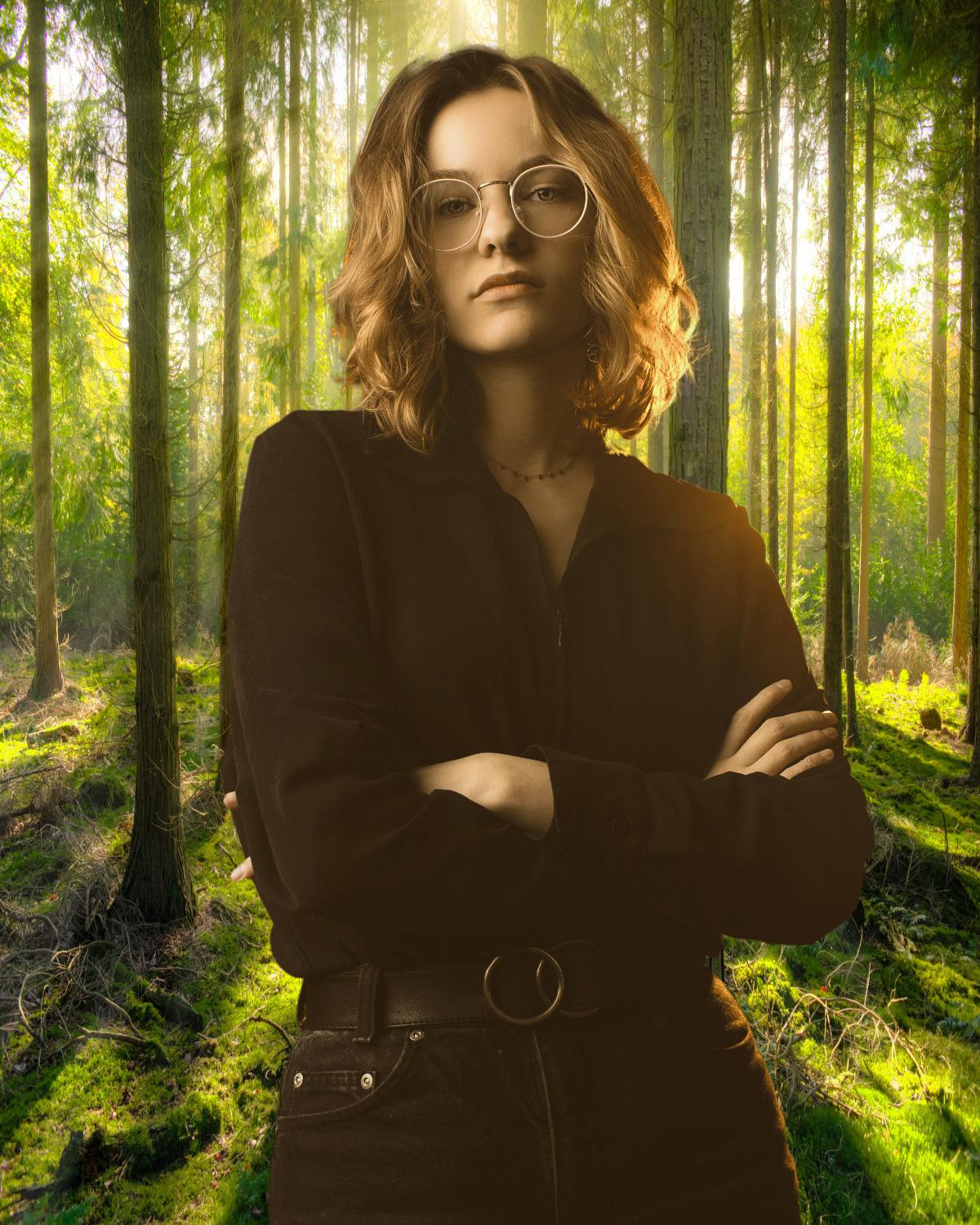}  

  \\
  DC \hspace{-1.2em} & GT \hspace{-1.2em} & $WCT^{2}$ \cite{yoo2019photorealistic} \hspace{-1.2em} & DIH \cite{tsai2017deep} \hspace{-1.2em} & $S^2$AM \cite{cun2020improving} \hspace{-1.2em} & DoveNet \cite{cong2020dovenet} \hspace{-1.2em} & SSH
  \\
\vspace{-1em}
\end{tabular}
\caption{\textbf{Comparison with the State-of-the-art methods.} The first column ``DC" represents the direct compositing results of foreground and background images, second column ``GT" represents the ground truth harmonized results annotated by human expert users. The rest five column shows the output of comparing methods and our proposed SSH methods. More visual results will be shown in the supplementary. Best viewed zoomed in.}
\label{fig:comparison}
\vspace{-1em}
\end{figure*}

\subsection{Dataset Preparation} \label{sec:dataset}
\subsubsection{Training Data and LUT Collection}
Since SSH requires only self-supervision (rather than the labeled object masks or harmonization ground truth from professionals user required by previous methods \cite{tsai2017deep,cun2020improving,cong2020dovenet}), we are able to collect a larger-scale unlabeled training set of images with diverse semantics, environmental lighting conditions and contents. Our unlabeled training set consists  of 81917 images from several datasets \cite{lin2014microsoft,bychkovsky2011learning,zhou2016evaluating} and the  Internet~\footnote{https://unsplash.com}\footnote{https://www.flickr.com/}, containing mountain, river, sky, general object, human portrait under diverse lighting conditions.  In addition, we also collect $100$ 3D color lookup tables from the Internet. 
We randomly select two LUTs from the collection in each training iteration, causing $100 \times 100$ possible combinations that enable diversity during training.

\subsubsection{Real-World Image Harmonization Benchmark}
Although there exist several testing datasets \cite{tsai2017deep,cong2020dovenet} for evaluating the harmonization performance, all of them are generated by perturbing the foreground objects with simple color-transfer methods \cite{reinhard2001color,xiao2006color,fecker2008histogram,pitie2007automated}.
Moreover, these methods \cite{cong2020dovenet,tsai2017deep} apply the same data synthesis procedure for both training stage and testing stage. For learning-based approaches, the evaluation results will be inevitably biased and cannot faithfully probe their real-world generalization. 

To bridge the gap between current evaluation protocols with real image harmonization demands, we propose a \textbf{Real}-world \textbf{H}ar\textbf{M}onization dataset for evaluation, named \textbf{RealHM}. However, collecting a well-annotated real-world testing dataset is not a trivial task, as several elements need to be adjusted together and the best result is beyond one image.
In particular, at least three main steps are required to generate high-quality harmonized composition, as follows: 1) A challenging compositing pair need to be picked where directly compositing these two images will cause severe visual disharmony. 2) A high-quality mask to cut out the foreground objects is needed, where the hard boundary and soft boundary (hair/fur region) are treated differently. 3) Matching the appearance (color,brightness,saturation,contrast) of the foreground with the background by using PhotoShop tools, where some particular local region needs further adjustment. A typical example is shown in Fig.~\ref{fig:adjustment}.

Finally, we collect 216 high-quality, high-resolution foreground/background pairs with corresponding harmonized outputs, where the foregrounds include both the human portraits and general objects and the backgrounds cover diverse environments such as mountain, river, buildings, sky and else.



\section{Experiments}
\subsection{Implementation}
SSH is first trained from scratch for 70
epochs with the learning rate of 2e-4, followed by another
30 epochs with the learning rate linearly decayed to 0. We adopt a scale jittering range of [256, 320] and then crop a $224 \times 224$ image in the training stage.
We use the Adam optimizer and the batch size is set to be 64. The whole training process takes 20 hours on 8 Nvidia 2080Ti GPUs. We implement it with PyTorch \cite{paszke2019pytorch} framework. We use $256 \times 256$ resolution for full-reference metrics
evaluation (PSNR, MSE, SSIM, and LPIPS) and use the
original resolution when showing the visual comparison for
better view. We follow~\cite{afifi2020deep} to generate high-resolution visual examples using color mapping function.

\subsection{Comparing with the State-of-the-art methods}
In this section we compare the performance of our proposed method SSH with the current state-of-the-art methods. We conduct both the quantitative and qualitative evaluation, including visual quality comparison, referenced image quality assessment (IQA) and human subjective test.

\begin{table*}[t]
\small
\begin{center}
\setlength{\tabcolsep}{10mm}{
\begin{tabular}{c|c|c|c|c} \toprule
Methods & PSNR $\uparrow$ & MSE$ \downarrow$ & SSIM $\uparrow$ & LPIPS $\downarrow$ \\
\specialrule{.1em}{.1em}{.1em}

DC & 25.91 & 409.54 & 0.9385 & 0.049 \\
\hline
$WCT^{2}$ \cite{yoo2019photorealistic} & 22.13 & 446.85 & 0.8559 & 0.096 \\
\hline
DIH \cite{tsai2017deep} & 23.96 & 433.52 & 0.8661 & 0.082\\
\hline
$S^2$AM \cite{cun2020improving} & 26.77 & 283.27 & 0.9366 & 0.053 \\
\hline
DoveNet \cite{cong2020dovenet} & 27.41 & 214.11 & 0.9416 & 0.044 \\
\hline
SSH & \textbf{27.91} & \textbf{206.85} & \textbf{0.9479} & \textbf{0.039} \\
\bottomrule
\end{tabular}}
\end{center}
\vspace{-1.5em}
\caption{\textbf{Comparing with the State-of-the-art methods}. We compare the the proposed method SSH with others with referenced-metric including PSNR, MSE, SSIM, and perceptual metrics. Here $\downarrow$ represents the lower the better and $\uparrow$ represents the higher the better. Our method outperforms previous methods under all these four metrics.}
\label{table:reference_metric}
\end{table*}

\subsubsection{Visual Quality Comparison}
We firstly compare our SSH framework with current state-of-the-art methods in terms of visual quality, shown in Fig.~\ref{fig:comparison}. The first column shows the results by directly compositing foreground and background images. The second column represents the annotated harmonized results retouched by the human expert user with professional editing skill. The third to seventh column shows the output generated by: $WCT^{2}$ \cite{yoo2019photorealistic}, DIH\cite{tsai2017deep}, $S^2$AM \cite{cun2020improving}, DoveNet\cite{cong2020dovenet}, and our proposed SSH methods. From the results in Fig.~\ref{fig:comparison}, photorealistic style transfer method \cite{yoo2019photorealistic} generates the worst results due to the foreground and background do not share similar layout, violating the prerequisite of its suitable scenarios. The results of DIH show subtle adjustment compared to the inputs since they only use one color transfer method to synthesize training data. The results of $S^2$AM and DoveNet either show incorrect color or disharmony contrast. In contrast, SSH successfully not only learns to extract correct color but also avoids contrast/brightness mismatching. More results will be shown in the supplementary.

\subsubsection{Full-referenced metrics}
\begin{figure}[t]
\centering
\resizebox{1\linewidth}{!}{
\begin{tabular}[!t]{cccc}
  \includegraphics[trim={0 0 0 0},clip,width=0.24\linewidth]{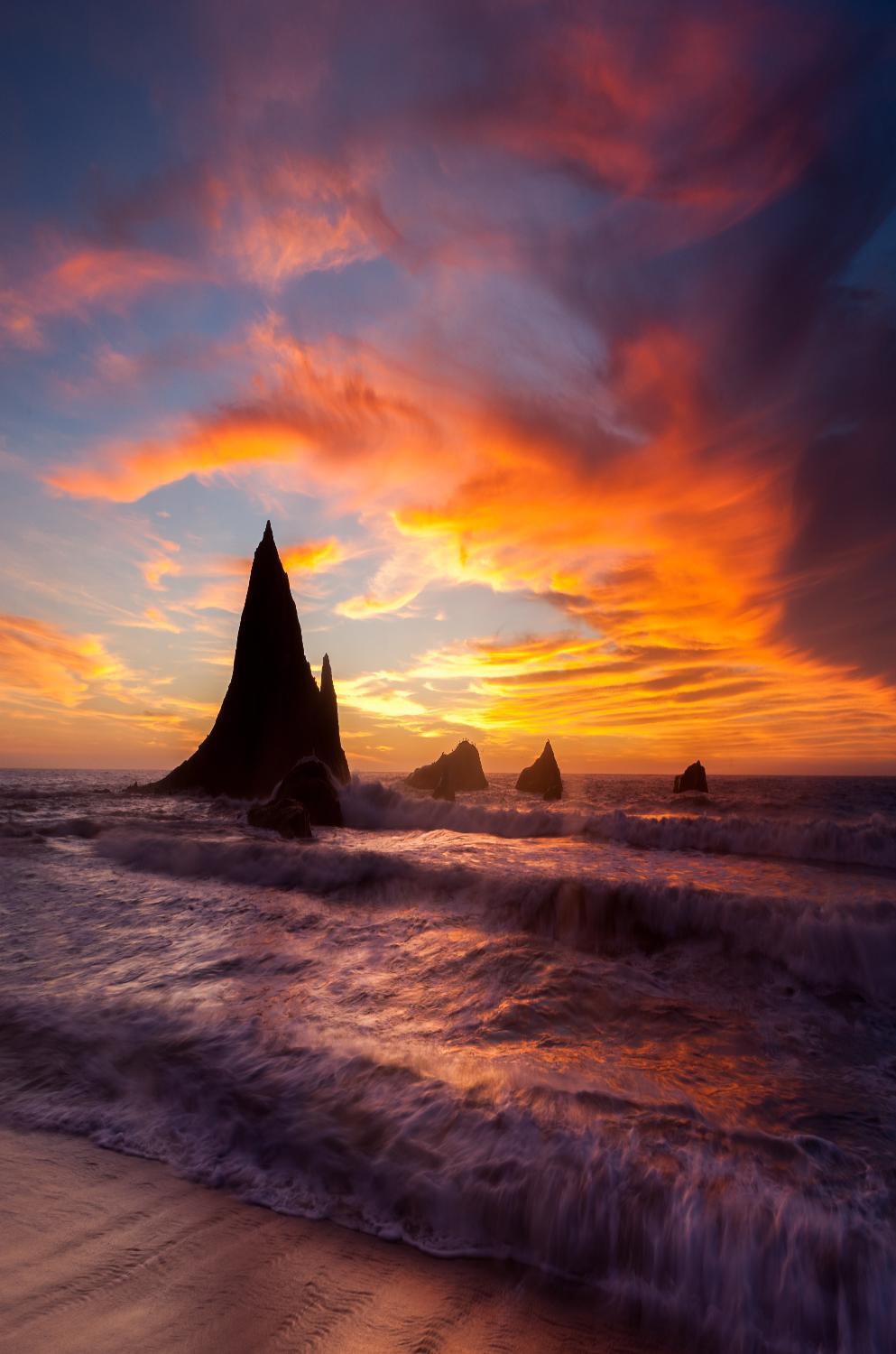} &  \hspace{-1.3em}
  \includegraphics[trim={0 0 0 0},clip,width=0.24\linewidth]{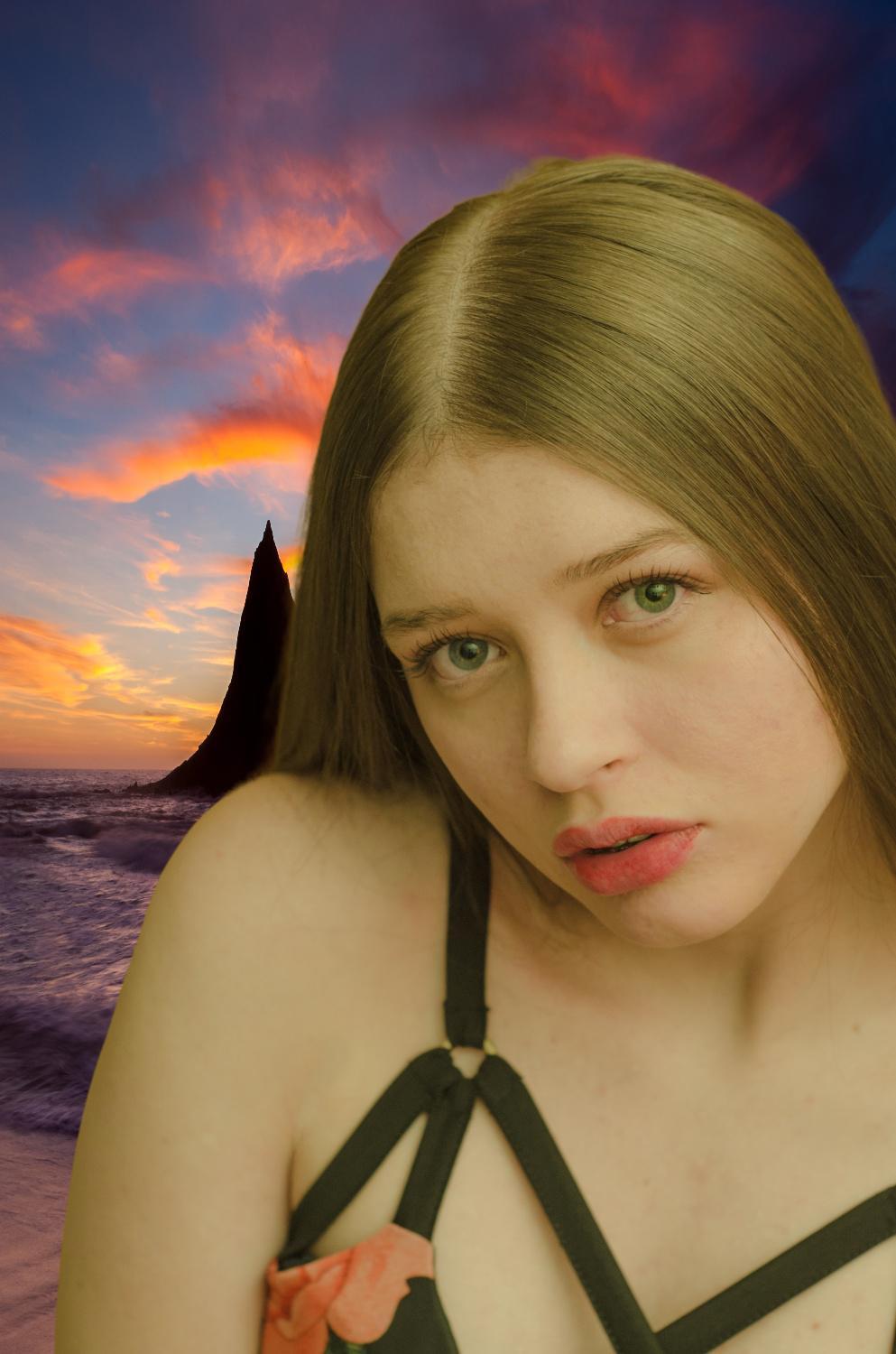}  &  \hspace{-1.3em}
  \includegraphics[trim={0 0 0 0},clip,width=0.24\linewidth]{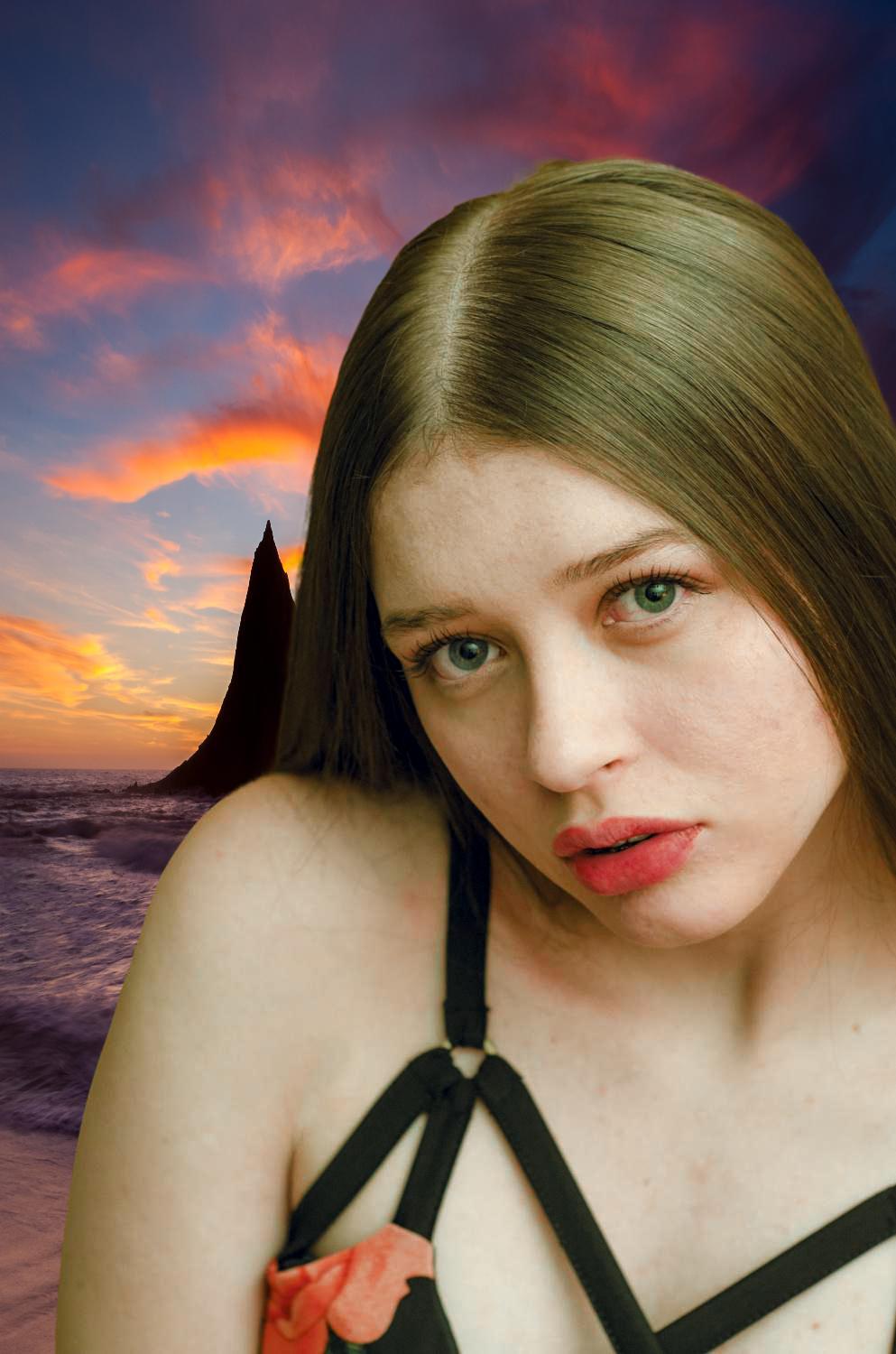} &  \hspace{-1.3em}
  \includegraphics[trim={0 0 0 0},clip,width=0.24\linewidth]{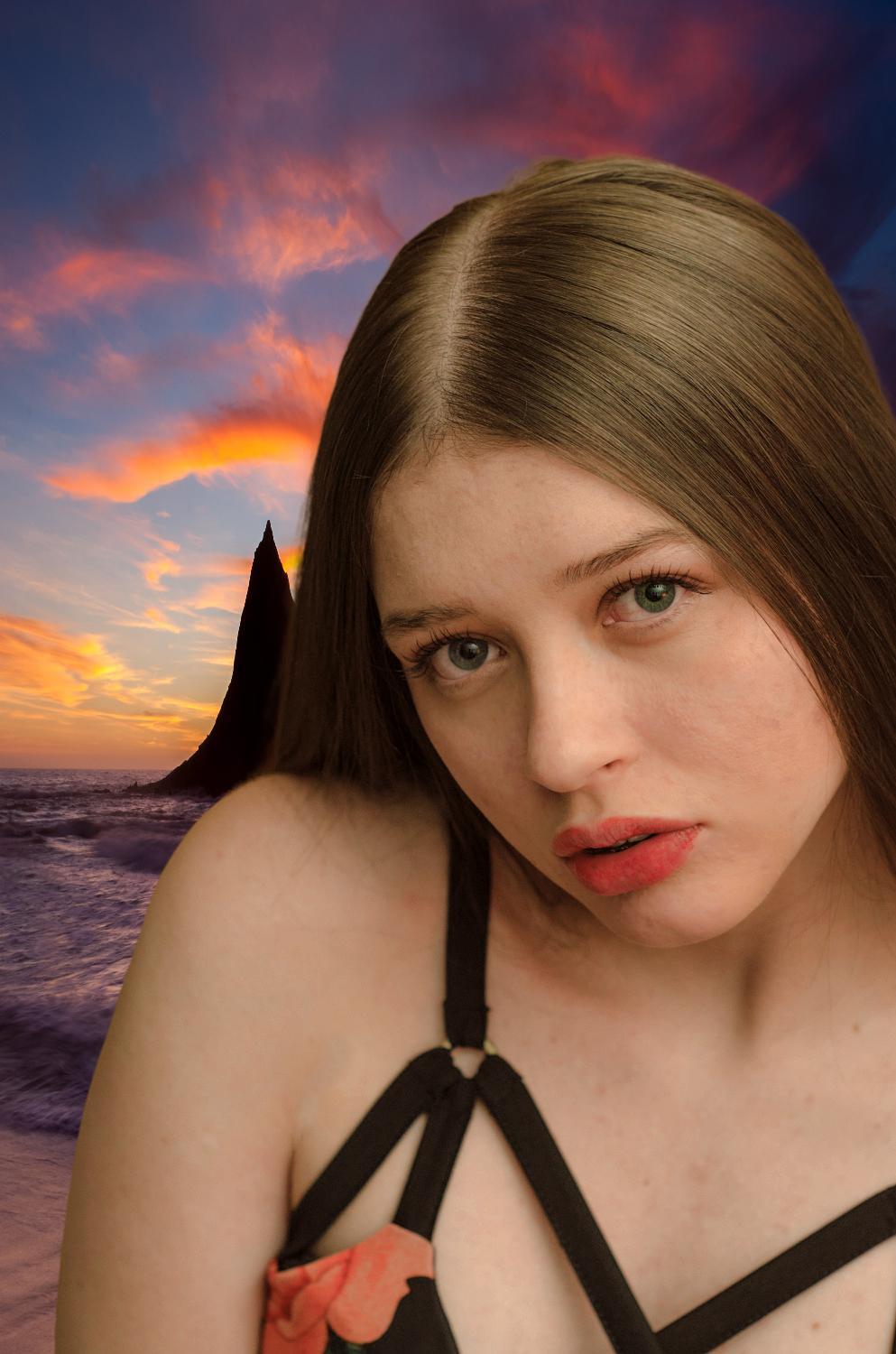} 
  \vspace{-0.3em}
  \\
  
  BG & \hspace{-1.3em}  DC  & \hspace{-1.2em} DoveNet \cite{cong2020dovenet}  & \hspace{-1.3em} SSH
\end{tabular}
}
\vspace{-1.em}
\caption{\textbf{Background occlusion problem.} ``BG" and ``DC" represent the background image and the direct compositing result, respectively. Since DoveNet~\cite{cong2020dovenet} only takes the direct compositing image as the input, it fails to capture the true appearance of the background when the foreground object is too large, while SSH successfully generates a harmonized output.}
\label{fig:back}
\vspace{-1em}
\end{figure}

We next evaluate the performance of these methods on our \textbf{RealHM} benchmark. Specifically, the human annotated harmonized output is set to be the ground truth label in each example, and the direct compositing results is considered as a baseline score here. Following previous work \cite{cun2020improving}, we adopt referenced metrics, including PSNR, Mean Squared Error (MSE), SSIM \cite{wang2004image}, and LPIPS \cite{zhang2018unreasonable}. The results are shown in Table \ref{table:reference_metric}.  To be more detailed, $WCT^{2}$~\cite{yoo2019photorealistic} shows the worst performance since its applicable scenarios requires the two images to share the similar layout (\eg, building-to-building). Besides, $S^2$AM \cite{cun2020improving} outperforms DIH \cite{tsai2017deep} due to its dual attention module. DoveNet \cite{cong2020dovenet} is slightly better than $S^2$AM thanks to its domain verification discriminator. However, these learning-based method all requires human annotated object mask. Different from those approaches, SSH reaches the best performance among all these four metrics without the necessity of any labels, demonstrating the effectiveness of the proposed self-supervised framework. We also include a typical example in Fig.~\ref{fig:back} to show that the back ground occlusion issue can be well addressed by the proposed SSH framework.

\begin{table}[t]
\begin{center}
\small
\setlength{\tabcolsep}{9mm}{
\begin{tabular}{c|c} \toprule
Methods & Score $\uparrow$ \\
\specialrule{.1em}{.1em}{.1em}

$WCT^{2}$ \cite{yoo2019photorealistic} & 0.821 \\
\hline
DIH \cite{tsai2017deep} & 1.201 \\
\hline
$S^2$AM \cite{cun2020improving} & 1.744 \\
\hline
DoveNet \cite{cong2020dovenet} & 1.256 \\
\hline
SSH & \textbf{2.295} \\
\bottomrule
\end{tabular}}
\end{center}
\vspace{-1em}
\caption{\textbf{Human Subjects Evaluation}. The higher score indicates the better result.}
\label{table:user_study}
\vspace{-1em}
\end{table}

\begin{table}[t]
\newcommand{\tabincell}[2]{\begin{tabular}{@{}#1@{}}#2\end{tabular}} 
\begin{center}
\small
\setlength{\tabcolsep}{1.6mm}{
\begin{tabular}{l|c|c|c|c} \toprule
Methods & PSNR $\uparrow$ & MSE $\downarrow$ & SSIM $\uparrow$ & LPIPS $\downarrow$ \\
\specialrule{.1em}{.1em}{.1em}
Single-Crops & 24.80 & 376.74 & 0.9203  & 0.069 \\
\hline
Multi-Crops (\textbf{ours}) & \textbf{27.91} & \textbf{206.85} & \textbf{0.9479} & \textbf{0.039} \\
\bottomrule
\end{tabular}}
\end{center}
\vspace{-1em}
\caption{\textbf{Evaluation of Content Augmentation}. The first row shows the results of SSH with single-cropping augmentation and second row shows the results of SSH with random multi-cropping augmentation (the proposed one).}
\label{table:ablate_content}
\end{table}

\newcommand{\tabincell}[2]{\begin{tabular}{@{}#1@{}}#2\end{tabular}} 
\begin{table}[t]
\begin{center}
\small
\setlength{\tabcolsep}{2mm}{
\begin{tabular}{c|c|c|c|c} \toprule
Methods & PSNR $\uparrow$ & MSE $\downarrow$ & SSIM $\uparrow$ & LPIPS $\downarrow$ \\
\specialrule{.1em}{.1em}{.1em}
color transfer & 27.01 & 311.96 & 0.9451 & 0.043 \\
\hline
Saturation & 26.68 & 337.11 & 0.9432 & .0460 \\
\hline
3D LUT \textbf{(ours)} & \textbf{27.91} & \textbf{206.85} & \textbf{0.9479} & \textbf{0.039} \\
\bottomrule
\end{tabular}}
\end{center}
\vspace{-1em}
\caption{\textbf{Evaluation of Appearance Augmentation}. We evaluate different ``appearance" augmentation strategy including color transfer, random saturation change, and our proposed 3D color lookup table (LUT) augmentation. }
\label{table:ablate_style}
\end{table}

\begin{table}[t]
\small
\begin{center}
\setlength{\tabcolsep}{1.5mm}{
\begin{tabular}{c|c|c|c|c} \toprule
Methods & PSNR $\uparrow$ & MSE $\downarrow$ & SSIM $\uparrow$ & LPIPS $\downarrow$ \\
\specialrule{.1em}{.1em}{.1em}
w/o recon loss. & 24.05 & 475.45 & 0.9095 & 0.0780\\
\hline
w/o disentangle loss & 24.94 & 372.21 & 0.9217 & 0.0643\\
\hline
SSH  & \textbf{27.91} & \textbf{206.85} & \textbf{0.9479} & \textbf{0.039} \\
\bottomrule
\end{tabular}}
\end{center}
\vspace{-1em}
\caption{\textbf{Evaluation of Loss Function Design Choices}.  The first row shows the results generated by SSH without reconstruction loss and the second row shows the results without disentanglement loss}
\label{table:ablate_loss}
\end{table}

\vspace{-0.5em}
\subsubsection{Human Subjects Evaluation}
We conduct a human subjective review to compare the performance of SSH with other methods. We randomly select 15 foreground and background pairs from the RealHM benchmark. Each image is first processed by five methods $WCT^{2}$~\cite{yoo2019photorealistic}, DIH~\cite{tsai2017deep}, $S^2$AM\cite{cun2020improving}, DoveNet\cite{cong2020dovenet}, and SSH), and then displayed on a screen for comparison. The human annotated ground truth is also provided as a reference. We then ask 50 subjects to independently score the visual quality, considering the following factors: 1) whether the image contains color/saturation disharmony; 2) whether the foreground and the background has different illumination; and 3) whether the composite images show texture distortions/artifacts. The score of visual quality ranges from 0 to 4 (worst to best quality). As shown in the Table \ref{table:user_study}, the photorealistic style transfer method $WCT^{2}$ \cite{yoo2019photorealistic} shows the worst performance since it is not suitable for image harmonization task, which is consistent with the observation from full reference evaluation. The learning-based image harmonization methods \cite{tsai2017deep,cun2020improving,cong2020dovenet} show comparably good performance, while the proposed method SSH achieves best score.


\subsection{Ablation Study} \label{sec:ablation}
\subsubsection{Effectiveness of Dual Data Augmentation}
As the superiority of SSN is built upon the strong dual data augmentation, we evaluate each component starting from the content augmentation and appearance augmentation. The content augmentation act as a crucial role in our framework as it generate two different crops of the same image, simulating the real scenarios where the foreground and background images are totally different. We ablate it by replacing the multi-cropping method with a single-cropping method and then make the content and reference network receive the same crop as the input. As shown  in Table \ref{table:ablate_content}, we observe that without adopting multi-cropping strategy, the performance largely degrades due to the pseudo label is exactly same as the input in the training stage, making the model easily minimizing the loss without actually learning the representation of the appearance.

Furthermore, we study the effectiveness of 3D LUT by comparing it with other appearance augmentation such as color transfer and random saturation adjustment. As shown in Table \ref{table:ablate_style}, the proposed 3D LUT augmentation strategy outperforms both the color transfer and random saturation change. This is because 3D LUT provides a stronger appearance change thanks to its diverse color mapping way. Also, 3D LUT enable local appearance change instead of a simple global translation.

\subsubsection{Evaluation of Loss function}
To study the effectiveness of reconstruction loss and disentanglement loss, we conducted the ablation experiments by removing them separately. As shown in Table \ref{table:ablate_loss}, either removing the reconstruction loss or the disentanglement loss will cause performance degradation, demonstrating the effectiveness of these loss. We also find that the disentanglement loss can help stabilize the performance our method, the visual examples will be shown in the supplementary.

  
  
  

\begin{figure}[t]
\centering   
\includegraphics[width=1\linewidth]{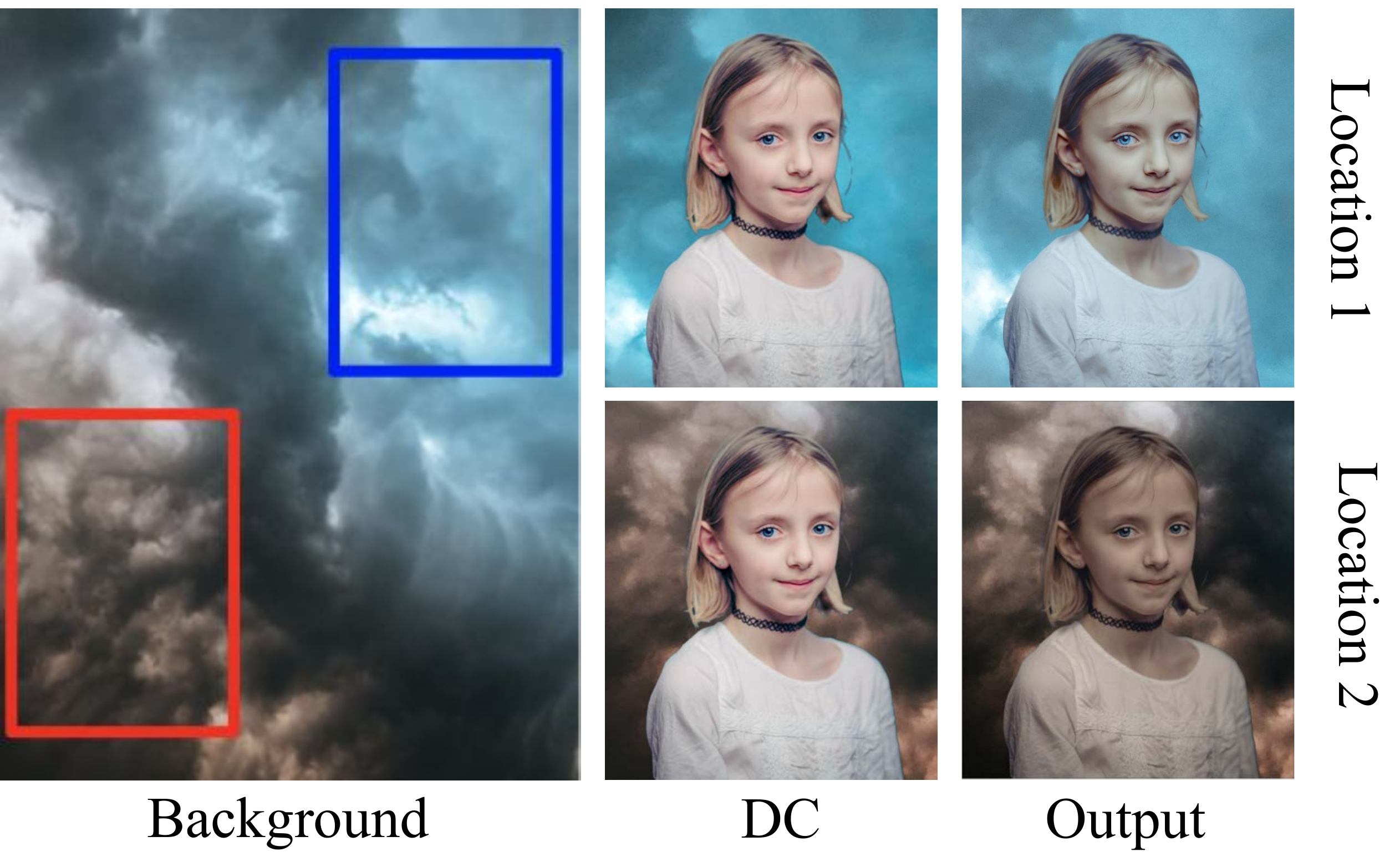}
\caption{\textbf{Locality-aware harmonization.} ``DC" denotes direct compositing results.} 
\label{fig:crop}
\vspace{-1em}
\end{figure}


\subsection{Locality-aware Harmonization}
Since the proposed self-supervised framework takes the full reference images as the inputs without annotated mask, the missing locality information brought by the annotated mask becomes a probable concern of current approach. In practice, we find that this issue can be well addressed by a proper cropping strategy during the inference stage. We take a spatially-variant colorful image as a typical background example, shown in Fig.~\ref{fig:crop}. The background contains two different appearance in different location, top right and bottom left. However, the appearance of the harmonized output is expected to be close to the region where the foreground object is placed. To avoid the harmonized output being affected by the misleading environment, we show that simply adopt the cropping method can well preserve the locality information. By separately considering the top right and bottom left cropping box as the background, the proposed methods successfully extract correct ``appearance" representation and generate reasonable and pleasing visual results, demonstrating the effectiveness of SSH in this challenging scenario.

\section{Conclusion}
We propose a self-supervised framework for image harmonization, named SSH. The proposed method does not require any human annotated labels in the training phase thus reduce the tedious effort of collecting a large-scale high-quality human annotated dataset retouched by professional users. Furthermore, we propose a dual data augmentation which include both the content data augmentation and appearance data augmentation to not only provide stable pseudo labels but also enrich the diversity of training data. Besides, we built a real harmonization benchmark that fills the gap in real testing scenarios. Our method outperforms all previous methods in a variety of metrics.

{\small
\bibliographystyle{ieee_fullname}
\bibliography{egbib}
}

\end{document}